\documentclass{article}

\usepackage{arxiv}

\usepackage[utf8]{inputenc} %
\usepackage[T1]{fontenc}    %
\usepackage{hyperref}       %
\usepackage{url}            %
\usepackage{booktabs}       %
\usepackage{amsfonts}       %
\usepackage{amsmath}
\usepackage[capitalise]{cleveref}
\usepackage{nicefrac}       %
\usepackage{microtype}      %
\usepackage{lipsum}		    %
\usepackage{graphicx}
\usepackage{float}
\usepackage{subcaption}
\usepackage{multicol}
\usepackage{doi}
\usepackage{comment}
\usepackage[square,numbers,sort&compress]{natbib}
\newcommand{\degree}{\ensuremath{^\circ}}
\newcommand{\scsim}{\textnormal{\raisebox{0.3ex}{$\scriptscriptstyle\sim$}}}

\title{Industrial Segment Anything -- a Case Study in\\Aircraft Manufacturing, Intralogistics,\\Maintenance, Repair, and Overhaul}

\author{ 
    \href{https://orcid.org/0000-0003-3892-1588}{\includegraphics[scale=0.06]{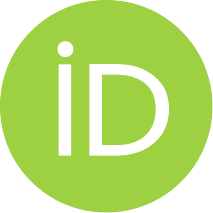}\hspace{1mm}Keno Moenck}\thanks{All authors are with \textit{Institute of Aircraft Production Technology}, \textit{Hamburg University of Technology}, \textit{Denickestraße 17}, \textit{21073 Hamburg}, \textit{Germany}.}\\
	\texttt{keno.moenck@tuhh.de} \\
    \And
    \href{https://orcid.org/0000-0002-5782-3468}{\includegraphics[scale=0.06]{orcid.pdf}\hspace{1mm}Arne Wendt$^{*}$}\\
	\texttt{arne.wendt@tuhh.de} \\
    \And
    \href{https://orcid.org/0000-0001-7385-8380}{\includegraphics[scale=0.06]{orcid.pdf}\hspace{1mm}Philipp Pr{\"u}nte$^{*}$}\\
	\texttt{philipp.pruente@tuhh.de} \\
    \And
    \href{https://orcid.org/0000-0003-3700-6425}{\includegraphics[scale=0.06]{orcid.pdf}\hspace{1mm}Julian Koch$^{*}$}\\
	\texttt{julian.koch@tuhh.de} \\
    \And
    \href{https://orcid.org/0009-0009-2528-5325}{\includegraphics[scale=0.06]{orcid.pdf}\hspace{1mm}Arne Sahrhage$^{*}$}\\
	\texttt{arne.sahrhage@tuhh.de} \\
    \And
    \href{https://orcid.org/0000-0001-5424-1502}{\includegraphics[scale=0.06]{orcid.pdf}\hspace{1mm}Johann Gierecker$^{*}$}\\
	\texttt{johann.gierecker@tuhh.de} \\
    \And
    \href{https://orcid.org/0000-0001-6488-5141}{\includegraphics[scale=0.06]{orcid.pdf}\hspace{1mm}Ole Schmedemann$^{*}$}\\
	\texttt{ole.schmedemann@tuhh.de} \\
    \And
    \href{https://orcid.org/0000-0003-4732-8364}{\includegraphics[scale=0.06]{orcid.pdf}\hspace{1mm}Falko K{\"a}hler$^{*}$}\\
	\texttt{f.kaehler@tuhh.de} \\
    \And
    \href{https://orcid.org/0000-0001-9918-349X}{\includegraphics[scale=0.06]{orcid.pdf}\hspace{1mm}Dirk Holst$^{*}$}\\
	\texttt{dirk.holst@tuhh.de} \\
    \And
    \href{https://orcid.org/0000-0002-5555-1139}{\includegraphics[scale=0.06]{orcid.pdf}\hspace{1mm}Martin Gomse$^{*}$}\\
	\texttt{martin.gomse@tuhh.de} \\
    \And
    \href{https://orcid.org/0000-0002-9616-3976}{\includegraphics[scale=0.06]{orcid.pdf}\hspace{1mm}Thorsten Sch{\"u}ppstuhl$^{*}$}\\
	\texttt{schueppstuhl@tuhh.de} \\
    \And
    \href{https://orcid.org/0000-0003-3912-1128}{\includegraphics[scale=0.06]{orcid.pdf}\hspace{1mm}Daniel Schoepflin}\thanks{Daniel Schoepflin is with \textit{Lufthansa Technik AG}, \textit{Weg beim Jäger 193}, \textit{22335 Hamburg}, \textit{Germany}.}\\
	\texttt{daniel.schoepflin@lht.dlh.de} \\
}

\renewcommand{\shorttitle}{Industrial Segment Anything}

\hypersetup{
pdftitle={Industrial Segment Anything},
pdfsubject={Preprint},
pdfauthor={K. Moenck},
pdfkeywords={machine vision, image segmentation, SAM, aircraft, production, manufacturing, logistics, MRO},
}

\begin{document}
\maketitle

\begin{abstract}

Deploying deep learning-based applications in specialized domains like the aircraft production industry typically suffers from the training data availability problem. Only a few datasets represent non-everyday objects, situations, and tasks.
Recent advantages in research around Vision Foundation Models (VFM) opened a new area of tasks and models with high generalization capabilities in non-semantic and semantic predictions. As recently demonstrated by the Segment Anything Project, exploiting VFM's zero-shot capabilities is a promising direction in tackling the boundaries spanned by data, context, and sensor variety. Although, investigating its application within specific domains is subject to ongoing research. 
This paper contributes here by surveying applications of the SAM in aircraft production-specific use cases. We include manufacturing, intralogistics, as well as maintenance, repair, and overhaul processes, also representing a variety of other neighboring industrial domains. Besides presenting the various use cases, we further discuss the injection of domain knowledge.

\end{abstract}


\setcounter{footnote}{0}%

\section{Introduction}
Computer or machine vision applications are fundamental components in enabling various industrial activities based on a digitized product or process \cite{naumannLiteratureReviewComputer2023, raiMachineLearningManufacturing2021, zhouComputerVisionTechniques2023,grosserIntegratedSolutions3d2012,sarkerDeepLearningComprehensive2021}. Especially within the aerospace production industry, shaped by small lot sizes, predominantly manual labor, and highly regulated products and processes, various use cases depend on computer vision tasks \cite{moenckDigitalTwinsAircraft2023, yasudaAircraftVisualInspection2022,giereckerConfigurationEnablementVision2022}, e.g., object recognition and tracking \cite{kahlerAIbasedEndpointDetection2023,busch_towards_2023,kheddarHumanoidRobotsAircraft2019}, classification \cite{schoepflinSmartMaterialDelivery2021,huLSTMRNNbasedDefectClassification2019,ruizDetectionClassificationAircraft2020,majiFineGrainedVisualClassification2013a}, segmentation \cite{dingVisualInspectionAircraft2022,zhaoMaskRCNNBased2021,moenck_towards_2022}, or anomaly detection \cite{kahlerAnomalyDetectionIndustrial2022, schmedemannDeepAnomalyDetection2022}. As in many industries \cite{omahonyDeepLearningVs2020}, deep learning-based applications usually outperform traditional computer vision techniques by higher accuracy and better generalization, which is necessary in dynamic and changing environments or processes. Pre-trained models based on publicly available datasets solve various real-world problems, but fine-tuning with custom data containing the specific object and sensor domain is usually necessary to obtain satisfying results. However, especially in distinct and unique domains, the training data availability problem is heavily pronounced \cite{janieschMachineLearningDeep2021,wuestMachineLearningManufacturing2016} since the few existing real-world datasets for vision use cases in industrial environments, e.g., \cite{bergmannMVTecADComprehensive2019, drostIntroducingMVTecITODD2017, lofflerEvaluationCriteriaInsideOut2018, rutinowskiSemiAutomatedComputerVision2023, willis_fusion_2021}, cover only a small number of objects and, based on annotations, only a few tasks can be trained. That is why recent research has been heavily focused on synthetic data encoding domain-specific knowledge \cite{schoepflinSyntheticTrainingData2021,dekhtiarDeepLearningBig2018,alexopoulosDigitalTwindrivenSupervised2020,manettasSyntheticDatasetsDeep2021} and domain adaptation / transfer learning \cite{bashkirovaVisDA2022Challenge2023, liDomainAdaptationYOLOv52023, zhangVisualInspectionSteel2021}, transferring a model from a source to a related target domain. Further approaches to reduce the domain-specific dataset sizes are one-shot / few-shot and zero-shot learning. Here, the model's pre-training primarily influences performance, so larger models and datasets directly increase it subject to the recently termed foundation models.

Foundation models trained on data at scale, in a self-/semi-supervised manner, shaped different aspects of artificial intelligence (AI) towards generic knowledge representations by not only being capable of applying to one task but a variety of downstream tasks, even novel ones that were not part of the pre-training stage through, e.g., fine-tuning and prompt engineering \cite{bommasaniOpportunitiesRisksFoundation2022}. In recent years, Natural Language Processing (NLP) scientists working on Large Language Models (LLM) made the most ground-breaking successes in this field, resulting in, e.g., BERT \cite{devlinBERTPretrainingDeep2019}, the GPT-n series \cite{ budzianowskiHelloItGPT22019, brownLanguageModelsAre2020, openaiGPT4TechnicalReport2023}, and LaMA \cite{touvronLLaMAOpenEfficient2023}. Following pure text en-/decoding, multi-modality was introduced through contrastive learning, as in the case of CLIP \cite{radfordLearningTransferableVisual2021} and ALIGN \cite{jiaScalingVisualVisionLanguage2021}, aligning learned visual and language representations for image-text retrieval, image generation, or even joining 2D/3D and textual inputs \cite{pengOpenScene3DScene2023, yangRegionPLCRegionalPointLanguage2023, zhangCLIPFO3DLearningFree2023}. Like Large Visual Models (LVM) followed LLMs as a result of transferring the attention and transformer concepts \cite{vaswaniAttentionAllYou2017} to vision tasks \cite{carionEndtoEndObjectDetection2020, dosovitskiyImageWorth16x162021,zhaiScalingVisionTransformers2022}, the vision community worked on Vision Foundation Models (VFM), resulting in the recently published, e.g., InternImage \cite{wangInternImageExploringLargeScale2023} and Segment Anything \cite{kirillovSegmentAnything2023} projects. Meta AI proposed the latter, including a promptable segmentation model and a dataset comprising 11 million images and 1 billion segmentation masks, making it currently the most ground-breaking VFM, even referred to as one of the "GPT-3 moments" in computer vision since the model learned some form of a general object concept, even on, during training, unseen objects and in ambiguous scenes \cite{jimfan[@drjimfan]ReadingMetaAISegmentAnything2023, zhangComprehensiveSurveySegment2023}.

The core strength of the SAM is its zero-shot capability, which raises the question of whether this capability transfers to different use cases in environments not included during training. Multiple recent publications deal with surveying use cases utilizing the SAM in different domains, e.g., medical image analysis \cite{kellener_utilizing_2023, lei_medlsam_2023, zhangHowSegmentAnything2023, chengSAMMedicalImages2023, dengSegmentAnythingModel2023, heComputerVisionBenchmarkSegmentAnything2023, huangSegmentAnythingModel2023, huBreastSAMStudySegment2023, huSkinSAMEmpoweringSkin2023, huWhenSAMMeets2023, iythasridharLungSegmentAnything2023, liPolypSAMTransferSAM2023, maSegmentAnythingMedical2023, mattjieZeroshotPerformanceSegment2023, mazurowskiSegmentAnythingModel2023, mohapatraSAMVsBET2023, qiuLearnableOphthalmologySAM2023, roySAMMDZeroshot2023, shiGeneralistVisionFoundation2023, wuMedicalSAMAdapter2023, zhouCanSAMSegment2023, semeraro_tomosam_2023}, geoinformation science / remote sensing \cite{chen_rsprompter_2023, shankar_segment_2023, renSegmentAnythingSpace2023, wangScalingupRemoteSensing2023, osco_segment_2023}, exogeology \cite{giannakis_deep_2023, julkaKnowledgeDistillationSegment2023}, construction \cite{ahmadiApplicationSegmentAnything2023}, material science \cite{semeraro_tomosam_2023}, biological imaging \cite{glatt_topological_2023}, sonar imaging \cite{wang_when_2023}, and agriculture \cite{williamsLeafOnlySAM2023, yangSAMPoultryScience2023}. For cross-domain surveys looking at the SAM, we refer readers to \cite{ji_segment_2023, zhangComprehensiveSurveySegment2023, zhangSurveySegmentAnything2023}.
In this work, we focus on use cases in producing industries based on examples in aircraft manufacturing, intralogistics, Maintenance, Repair, and Overhaul (MRO), to contribute real-world ideas and concepts of utilizing the SAM. To the best of our knowledge, we are the first to assess the use of the SAM in these domains, applied to different use cases -- always without claiming any completeness.

The remainder of this work is structured as follows: \cref{sec:sam} introduces the Segment Anything Model (SAM) in more detail and presents existing derivatives. Subsequently, we introduce the aircraft domain in \cref{sec:vision_aero} and continue to elaborate on use cases in \cref{sec:use_cases_manufacturing,sec:use_cases_logistics,sec:use_cases_mro} for manufacturing, intralogistics, and MRO, respectively. We summarize the use cases in \cref{sec:summary} and discuss the SAM’s domain robustness. Eventually, we conclude in \cref{sec:discss_outlook}. We will dynamically update this preprint to reflect the progress around the SAM and other VFMs but also applications and use cases in the depicted domain.

\section{Segment Anything Model (SAM)}\label{sec:sam}

The Segment Anything project introduced three novel components: a promptable 2D segmentation task, the promptable model that partitions an image in semantically meaningful regions, and the SA-1b dataset with 11 million images and 1+ billion masks \cite{kirillovSegmentAnything2023}.

\begin{figure}[t]
    \centering
    \includegraphics[width=1.0\linewidth]{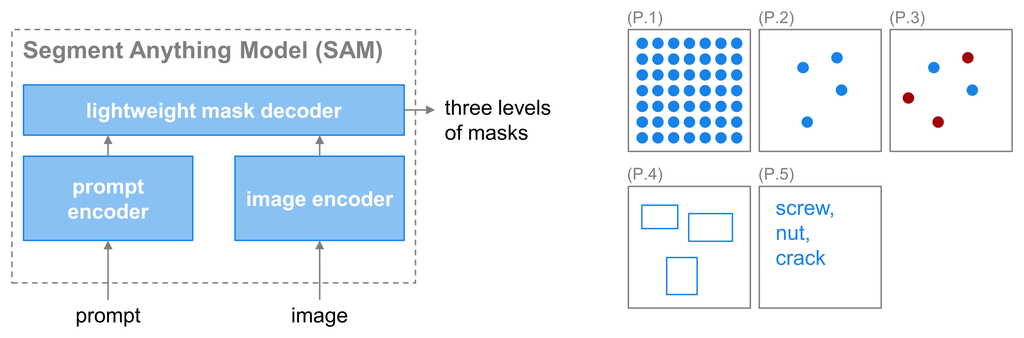}
    \caption{Overview of the Segment Anything Model (SAM) (left) and different types of sparse and dense prompts that we utilize in this work (right): regular grid sampling (\textbf{P.1}), single foreground point querying (\textbf{P.2}), fore- / background point querying (\textbf{P.3}), bounding box prompting (\textbf{P.4}), text prompting (\textbf{P.5}).}
    \label{fig:sam_prompts}
\end{figure}

The model comprises a prompt and image encoder whose outputs are combined in the lightweight mask decoder outputting three levels of masks per prompt and a confidence score, a predicted Intersection over Union (IoU) for each mask (s. \cref{fig:sam_prompts} left side). An MAE pre-trained Vision Transformer (ViT) is used as an image encoder. Since the image and prompt encoder are detached, an image must only run once through the encoder. The image embedding can then be reused for different prompts. So, the computationally expensive operation in which a GPU resource is suggestive is separated. The prompt encoding / decoding can even take place in a web browser using a CPU. That enables many novel use cases, e.g., interactive prompting in an Augmented Reality (AR) application on a mobile device.
The SAM outputs three levels of masks for each prompt that differ in size and try to represent three levels of partitions of an object, namely whole / part / sub-part. The SAM's authors state that this partition is sufficient for most cases. For selecting one mask out of multiple overlapping masks, non-maximum suppression (NMS) can be utilized, as we will do in this work. We observed that the SAM-predicted IoU and NMS can mostly remove ambiguous mask fragments.

The SAM's accompanying, reported to date largest known, segmentation dataset SA-1b was iteratively created in a three-staged process including model-assisted, semi-automatic, and fully automated annotation. Only masks were collected during the manual labeling process, and any semantics was discarded. The aim was to collect masks for prominent objects and not to gather masks for all visible objects and all their details in one image. Therefore, the labeling was constrained by time for each image. Retraining the SAM with newly collected data and scaling up the ViT resulted in an astonishing total of 1.1B masks, where $99.1\%$ were generated automatically.

In this study, we utilize the following prompt taxonomy (\textbf{P.1-5}, s. \cref{fig:sam_prompts} right side) while also processing ambiguous and overlapping masks by NMS. The automatic generation of masks is achieved through regular sampling of query points on a grid (\textbf{P.1}). We also utilize single foreground point querying (\textbf{P.2}), where masks for each point are prompted. Alternatively, single prompting based on single / multiple foreground and background points (\textbf{P.3}) is encouraged, especially to fine-tune ambiguous object parts manually. Finally, we utilize bounding box prompting (\textbf{P.4}) but also text prompting (\textbf{P.5}) based on GroundedSAM \cite{noauthor_grounded-segment-anything_2023}. The usage of other SAM derivates is mentioned accordingly. If not stated differently, we use the ViT-H image encoder and the default parameter set of the automatic segmentation module in the SAM Python implementation\footnote{\url{https://github.com/facebookresearch/segment-anything}}.

\subsection{Related Work}\label{sec:related_work}

Even though the SAM was only recently introduced, already proposed improvements and extensions underline the great interest of the vision community and the SAM's promising capabilities, e.g., HQ-SAM adds additional tokens and global-local feature fusion to increase the quality of the segmentation masks \cite{ke_segment_2023}, MobileSAM accelerates the image encoding by employing a lightweight image encoder \cite{zhang_faster_2023}.
A few authors propose to use the SAM in weakly-supervised learning to extend sparsely-annotated data by generating pseudo labels \cite{heWeaklySupervisedConcealedObject2023,jiangSegmentAnythingGood2023,sunAlternativeWSSSEmpirical2023,cui_all--sam_2023} or enhancing pseudo labels \cite{chenSegmentAnythingModel2023}. The SAM can generate pixel-level annotations for only image-level annotated data, which definitely decreases labeling costs.
Other related work uses the SAM in conjunction with further models, e.g., BLIP \cite{li_blip_2022} and Point-E \cite{nichol_point-e_2022} to enable a 3D reconstruction pipeline \cite{shenAnything3DSingleviewAnything2023} from single images or in conjunction with an image captioner to interactively, prompt-based caption images \cite{wangCaptionAnythingInteractive2023}.
Adding semantics to SAM-generated masks through close-set semantic segmentors or open-vocabulary classifiers is of interest as in \cite{chen2023semantic}. The other way around, the SAM can generate masks for objects in bounding boxes generated by a zero-shot object detector, e.g., Grounding DINO \cite{liu_grounding_2023}, as presented in \cite{noauthor_grounded-segment-anything_2023}. \cite{cao_segment_2023} uses the same approach to segment anomalies by incorporating domain expert knowledge to prompt the models. \cite{yan2023count} assembles CLIP and the SAM to count objects. 

Further examples utilizing the SAM are concealed object segmentation \cite{heWeaklySupervisedConcealedObject2023}, removing shadows in images \cite{jieWhenSAMMeets2023, wangDetectAnyShadow2023, zhang_sam-helps-shadowwhen_2023}, matting \cite{li_matting_2023, yao_matte_2023}, inpainting \cite{wangInpaintNeRF360TextGuided3D2023,yuInpaintAnythingSegment2023}, or general image editing \cite{jiangRestoreAnythingPipeline2023}, LiDAR to 2D camera calibration \cite{luo_calib-anything_2023}, video super-resolution \cite{luCanSAMBoost2023}, object tracking or segmentation in videos \cite{chengSegmentTrackAnything2023, yangTrackAnythingSegment2023, zhangUVOSAMMaskfreeParadigm2023, zhouDSECMOSSegmentAny2023, rajic_segment_2023, li_refsam_2023}, object counting \cite{yan2023count, maCanSAMCount2023a, shi_training-free_2023}, eye-tracking supported labeling of images \cite{wangGazeSAMWhatYou2023}, and open-world object detection \cite{he_usd_2023}.

Fine-tuning large models with small datasets and fewer computational resources is subject to current research. One strategy is inserting several parameter-efficient adapters into the transformer architecture where pre-trained weights are frozen during fine-tuning, e.g., Low-Rank Adaptation (LoRA) \cite{hu_lora_2021}. \cite{wuMedicalSAMAdapter2023} and \cite{zhangCustomizedSegmentAnything2023} already show this approach with promising results in the medical image domain. \cite{zhangPersonalizeSegmentAnything2023} adds externally trained weights to aggregate SAM-generated output masks adaptively. \cite{shaharabany_autosam_2023} replaces the SAM's prompt encoder with a custom one to fine-tune the model. \cite{liuMatcherSegmentAnything2023} show a complete training-free framework for solving few-shot segmentation tasks based on bidirectional matching between a sample and reference images.
Learning-based fine-tuning strategies like LoRA require a set of annotated data again, while the other mentioned few-shot strategies especially claim to increase the mask's robustness. 

Related works show various tasks in which the SAM’s or, in general, VFM’s capabilities can be exploited. On the one hand, this survey focuses on the zero-shot capabilities in the depicted domain; on the other hand, we elaborate on machine vision pipelines in which a VFM like the SAM can play a novel, significant role. Besides focusing on the aircraft production domain, we try to always also discuss the ideas in the context of neighboring industrial environments.

\begin{figure}[t]
    \centering
    \includegraphics[width=\linewidth]{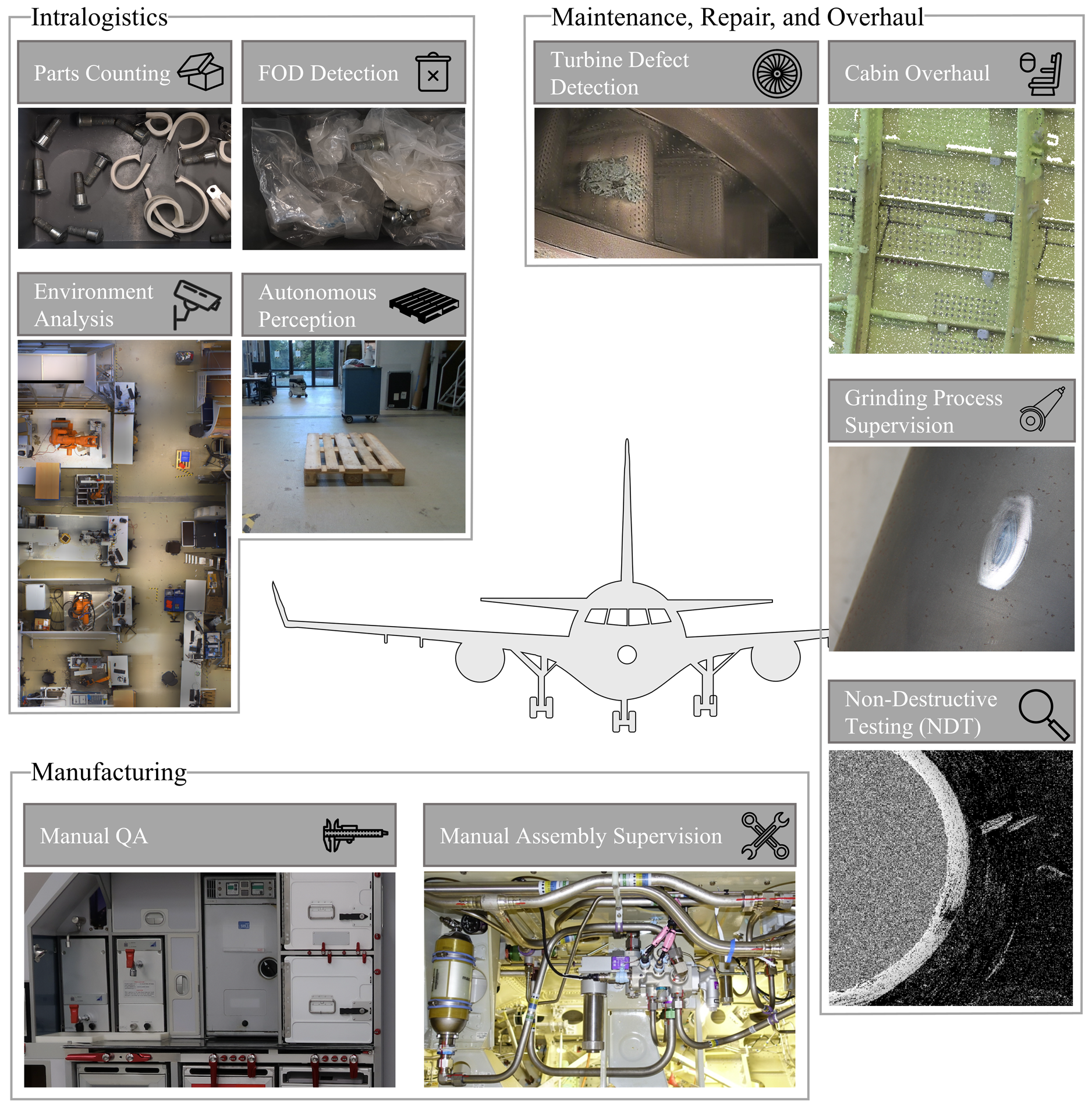}
    \caption{This work considers several use cases in manufacturing (s. \cref{sec:use_cases_manufacturing}), intralogistics (s. \cref{sec:use_cases_logistics}), and MRO (s. \cref{sec:use_cases_mro}).}
    \label{fig:ConsideredUseCases}
\end{figure}

\section{Applications in the Aircraft Industry}\label{sec:vision_aero}

With historically grown manufacturing environments, highly-individual MRO processes, and complex and simultaneously large-scale (sub-)assemblies, the aircraft industry has been a challenging domain for digitization and automation. Commercial aircraft manufacturing, with an overall low four-figure volume per year \cite{noauthor_recovery_2023}, is characterized by a close to lot size one manufacturing \cite{moenckDigitalTwinsAircraft2023}. Therefore, existing machine vision applications are highly specialized and, due to the costly development and software adaption processes, only found in a few processes, sometimes even discarded shortly after introduction if a part or process changes. Instead of engineering, applied research is dominating use cases and examples.

Highly individual products and processes, as in the domain of aircraft production, are not represented in publicly available datasets. Thus, current off-the-shelf trained models do not provide context-accurate output, as the semantics of the domain have not been trained. In contrast, the SAM's abstract and generalized yet profound capabilities to differentiate between \textit{any} objects / parts in a scene are promising. Exploiting these abilities, three strategies are possible considering the non-existence of domain knowledge:

\begin{enumerate}
    \item \textbf{Semantic-less segmentation:} Use the SAM in an application in which no further semantics is necessary to provide benefit, so no domain knowledge is required. Here, masks separating meaningful semantic regions are sufficient.
    \item \textbf{Embed} the SAM as one step of a pipeline so domain knowledge and context are provided in a different application stack.
    \item \textbf{Fine-tune} the SAM based on data encoding some form of domain knowledge.
\end{enumerate}

In the scope of this work, we will introduce vision application targets that fall in the first two approaches and mention scenarios for the third (s. \cref{fig:ConsideredUseCases} for a graphical overview).

Based on regulators, the aircraft industry is divided into four different types of organizations: design, production, maintenance, and continuing airworthiness management. After the initial production, the overall objective of each of these is to keep an aircraft’s airworthiness at all times while continuing airworthiness management is the contracting management body. Since there is rarely any activity in the initial lifecycle phase, the aircraft‘s and component‘s design phase, which depends on machine vision, fields of applications are within manufacturing and maintenance, repair, and overhaul -- short MRO. Besides, in discrete unit manufacturing, especially with the aircraft's millions of parts, the handling of these plays a particular role, where transportation falls under intralogistics, which we will treat as a distinct use case topic.

\section{Manufacturing}\label{sec:use_cases_manufacturing}
Vision systems that process optical acquired 2D images in manufacturing are, following \cite{hornberg_handbook_2017}, composed of one or more of the following tasks: code recognition, object or position recognition, completeness, shape / dimension check, or quantitative or qualitative inspection.

Within code recognition, objects or assets are identified through explicit markings on these, e.g., one-dimensional Barcodes, two-dimensional QR-Codes, or any other custom code. We also include Optical Character Recognition (OCR) in this category. This task is basically solved for high frequencies using traditional or off-the-shelf models – no necessity for applying a VFM.

In contrast, segmentation can be part of recognizing an object, its position, or its state. Here, the SAM may act in pre-processing an image, e.g., based on querying foreground and background points (P.3), or in post-processing by, e.g., generating pixel-level masks subsequent to a state-of-the-art object detector. Besides, one can utilize, e.g., the SAM‘s robustness against concealed or camouflaged objects \cite{ji_segment_2023}, which may outperform other (traditional) methods.

Completeness checks refer to observing the state of an assembly process, e.g., checking at a given point in time if all parts are in the correct position and present. Especially the aircraft industry-typical assembly conditions like high-quality standards, time-parallel operations within a large workspace, and use of numerous, similar parts introduce challenges for vision applications here \cite{gierecker_automated_2023}. Undetected errors during assembly pose safety risks and can impact subsequent production. In this context, we further discuss the application of the SAM in \cref{sec:assembly_progress_monitoring} on a specific example.

Further, objects' shape and dimension check applications can be assisted through the SAM in a pre-processing segmentation stage. Since here, some specific Geometric Dimensioning and Tolerancing (GD\&T) post-processing logic is the core component, we omit to provide an example and let this open for future work.

Finally, quantitative and qualitative inspections of surfaces as part of Quality Assurance (QA) are either done manually or (semi-)automatically. The current practices in aircraft QA heavily relies on manual procedures and document-based quality control, which can be time-consuming, error-prone, and delay information feedback. Assisting a manual task through a vision system ensures reliability and repeatability. Especially the newly introduced promptable two-dimensional segmentation task is promising here since it mimics the human behavior of pointing at \textit{things}. We will discuss this in more detail within a manufacturing use case in \cref{sec:manual_quality_control} and an MRO use case in \cref{sec:mro_visual_inspection}.

\subsection{Assembly Progress Monitoring and Inspection}\label{sec:assembly_progress_monitoring}

During aircraft manufacturing, assembly inspection plays a vital role in quality control by identifying misassemblies, including incorrect assembly positions, wrong part types, and missing parts. Currently, aircraft assembly relies on manual procedures and document-based quality control. Information feedback about performed assembly tasks or emerging problems is crucial for process control but is at most done by the worker and therefore delayed and prone to error. Furthermore, the presence of undetected errors can pose significant safety risks and impact subsequent production stages. Either assisted or automated in-progress monitoring or an end-of-line control using a sensor-based solution can help to overcome the mentioned problems.

\begin{figure}[t]
    \centering
    \begin{subfigure}[t]{\linewidth}
        \centering
        \begin{subfigure}[t]{0.495\linewidth}
            \caption{Progress monitoring: Observation of part presence}
            \includegraphics[width=\linewidth]{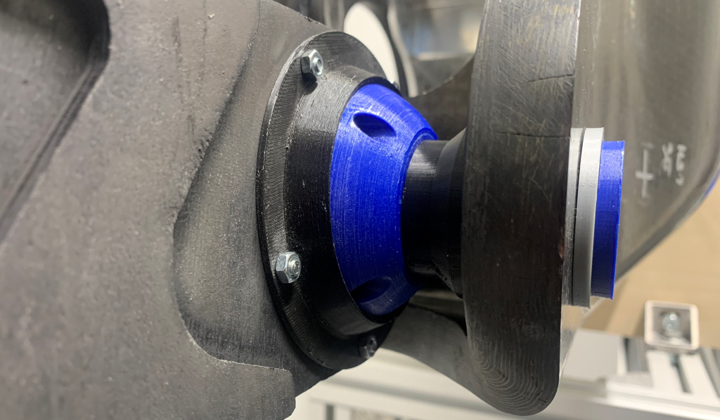}
            \includegraphics[width=\linewidth]{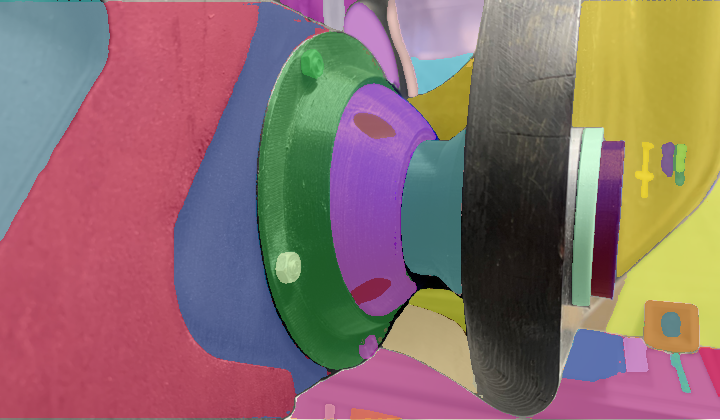}
        \end{subfigure}
        \begin{subfigure}[t]{0.495\linewidth}
            \caption{Final inspection: Assisted annotation and documentation}
            \includegraphics[width=\linewidth]{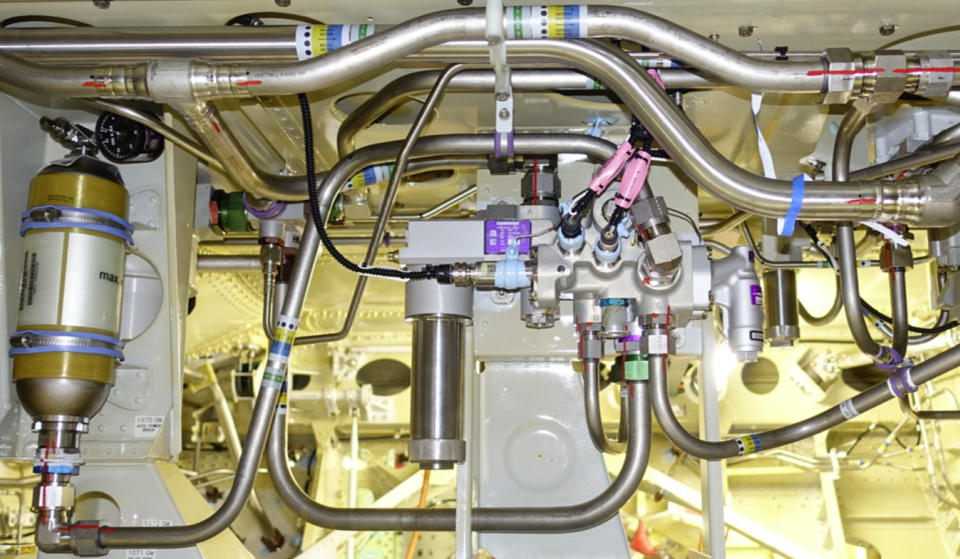}
            \includegraphics[width=\linewidth]{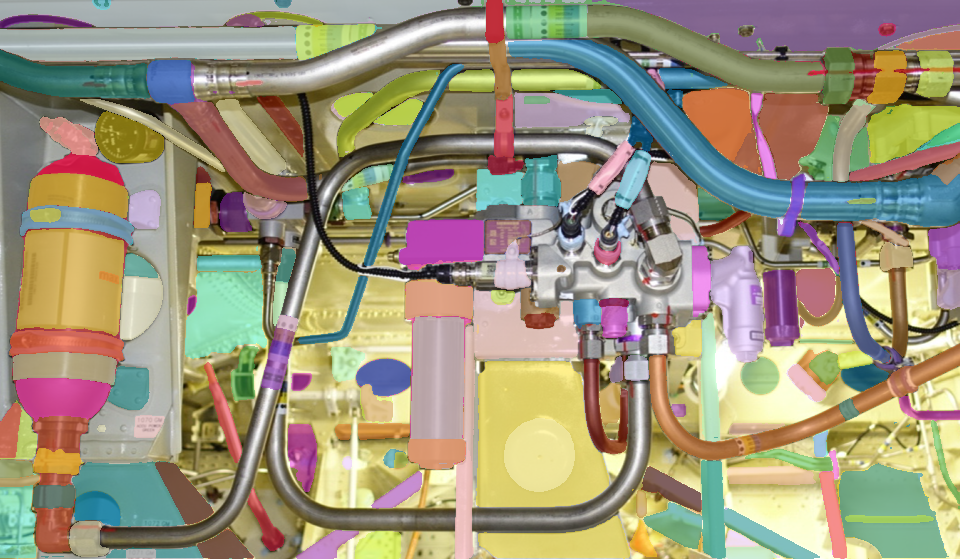}
        \end{subfigure}
    \end{subfigure}
    \caption{Application of the SAM on aircraft-typical assemblies: Segmentation of parts in articulated joints for the derivation of status information (a). Automatic (\textbf{P.1}) generation of segmentation masks in complex assemblies (landing gear bay of an Airbus A320 \cite{neumaier_fully_2022}) assists the user during annotation and documentation in the final assembly inspection (b).}
    \label{fig:AssemblySamples}
\end{figure}

Optical sensor-based assembly supervision can be a continuous or event-based method to verify the correctness and completeness of tasks by observing the assembly’s shape \cite{gierecker_assembly_2021}. Depending on the level of detail, part detection, recognition, localization, and classification are carried out every time a new image is available. Therefore, the different parts must be detected and segmented within the image frame. Especially within articulated assemblies, the SAM can provide advantages by easily generating masks for different parts regardless of their orientation, thus forming the basis for further processing steps. 

Another possible application of the SAM is within the final inspection of completed assemblies. As, e.g., in the aircraft domain, the numerous and various occurrences of similar, texture-less parts like brackets or pipes form a difficult environment for manual visual inspection. \Cref{fig:AssemblySamples} (b) shows the pipe installation with segmented parts in an Airbus’ main landing gear bay environment. An assistance tool based on the SAM can produce semantic-less segmentation masks, thus, the distinction of the different components within the complex environments for the user. That has the potential to simplify and speed up the documentation process by requiring the user to only perform the inspection using pre-selected masks and annotate the image material based on this.

\subsection{Manual Visual Inspections in Quality Control}\label{sec:manual_quality_control}

\begin{figure}[t]
    \centering
    \includegraphics[width=\linewidth]{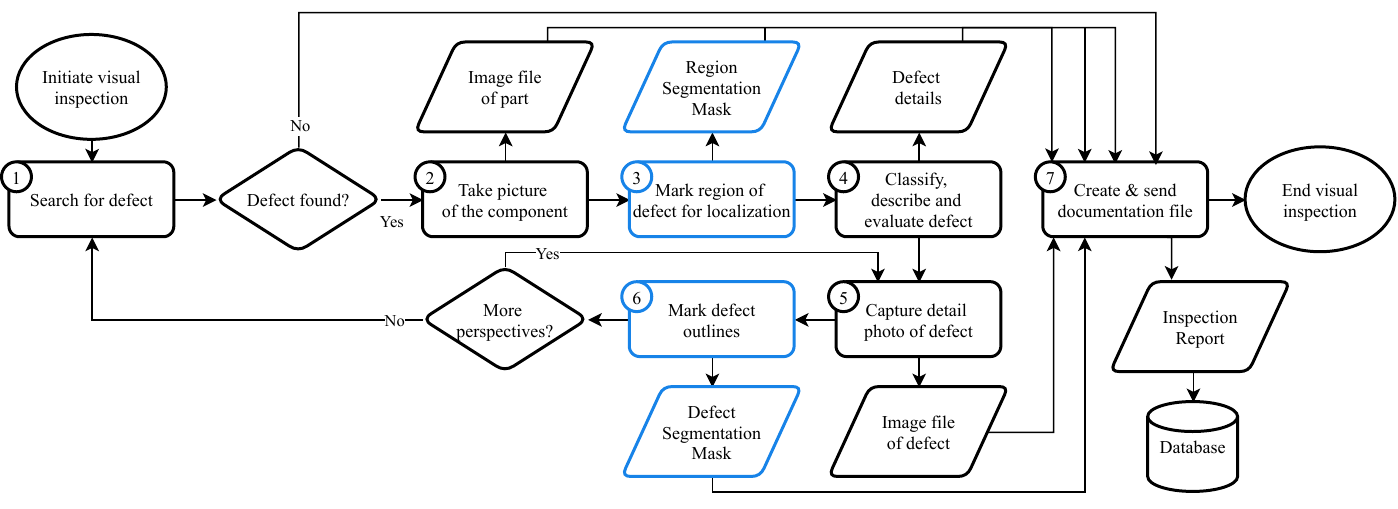}
    \caption{Flow chart of a manual visual inspection executed with a web application on a tablet (blue: potential processes to be supported by the SAM).}
    \label{fig:FlowDigram_QA}
\end{figure}

In addition to controlling the correct assembly of components, as mentioned in the section before, assessing the quality of the resulting intermediate and end products is a typical task. Along the value chain of components, the results of manufacturing and assembly processes are checked with the aid of quality characteristics within an inspection plan in order to detect defects at an early stage and reduce quality costs (e.g., for rework). For components with a large number of variants, such as aircraft cabin monuments, the number of manual process steps is especially high \cite{kalscheuer_reducing_2023}. Due to their non-value-adding characteristic, it is of economic interest to design quality processes as efficiently as possible, yet not influencing the information content of the process results \cite{rozanec_towards_2022}.

In this context, we consider a manual visual inspection of aircraft cabin monuments as our use case for this section which is facilitated by an application running on a tablet. The quality control aims to detect and document surface defects on the Devices Under Test (DUT) through a combination of visual examination, photo capturing, segmentation, and defect description.
The following paragraph provides a short description of each process step as depicted in \cref{fig:FlowDigram_QA}.

After initiating the process, the inspector commences by scrutinizing the surface of the DUT for visible defects or imperfections (1). Upon identifying a defect, the inspector employs the tablet's camera to capture a photo of the specific area of the defect on the DUT (2). This captured photo serves as the input for the tablet application's canvas on which the location of the defect can be marked with various segmentation tools (3). These tools include options for drawing a Region Of Interest (ROI) with rectangles, circles, or polygons. The inspector describes the defect in detail once the region of interest is defined (4). That involves utilizing a pre-defined classification system (e.g., dents, scratches), input fields to articulate further observations and a severity evaluation on a scale from light to heavy. To further assess the defect comprehensively, the inspector can capture one or more detailed photos (close-ups) from different perspectives using the tablet's camera (5). The image capturing of each photo is followed by an image segmentation process to mark the outlines of each defect (6) accurately. Within this process, the same segmentation tools are offered to the user as in process (3). When all necessary perspectives of the defect are captured, the inspection process enters a loop, returning to the first step. If more defects are present on the DUT, the inspector repeats the process described beforehand; if not, the inspection concludes with creating and sending the inspection report (7). This report incorporates the captured photos with segmentation masks, defect classification, textual descriptions, and severity evaluations.

The blue-marked steps in \cref{fig:FlowDigram_QA} are potential applications for implementations of the SAM.
We further investigated the potential by exploiting the SAM’s capabilities in mask generation on images usually taken during the visual inspection of cabin components. Instead of manually generating masks using the web application that had been demonstrated to be time-consuming, we utilized the SAM.

\begin{figure}[t]
    \centering
    \begin{subfigure}[b]{\linewidth}
        \centering
        \begin{subfigure}[b]{0.495\linewidth}
            \caption{Input data \cite{noauthor_diehl_2018}}
            \includegraphics[width=\linewidth]{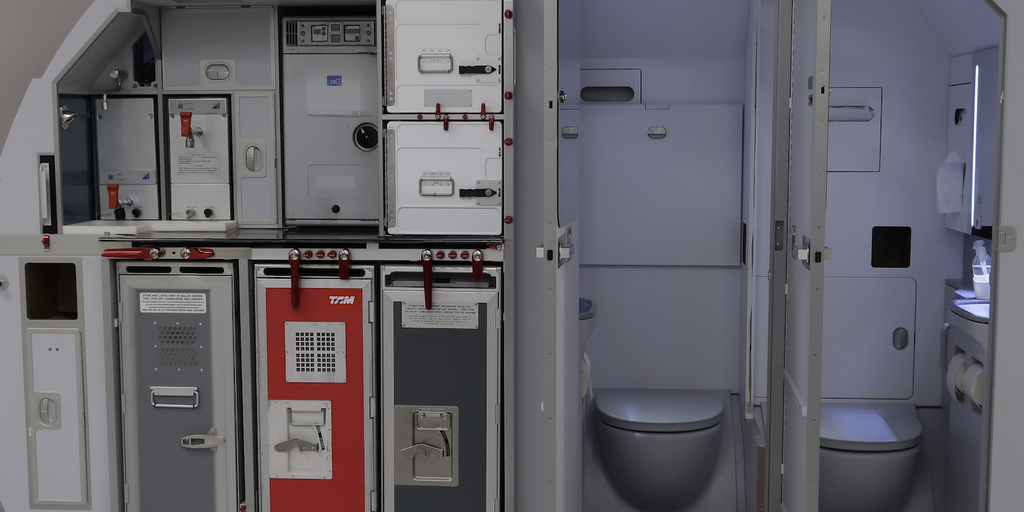}
        \end{subfigure}
        \begin{subfigure}[b]{0.245\linewidth}
            \caption{Shelf}
            \includegraphics[width=\linewidth]{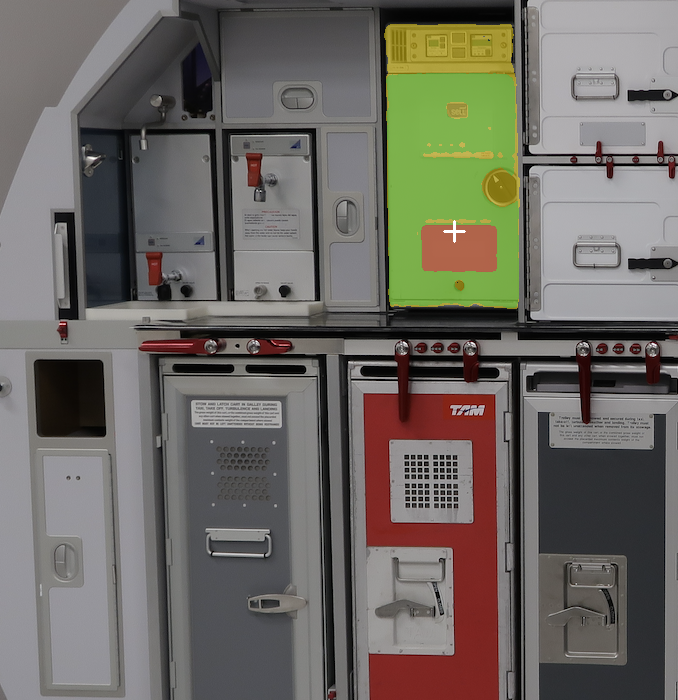}
        \end{subfigure}
        \begin{subfigure}[b]{0.245\linewidth}
            \caption{Toilet}
            \includegraphics[width=\linewidth]{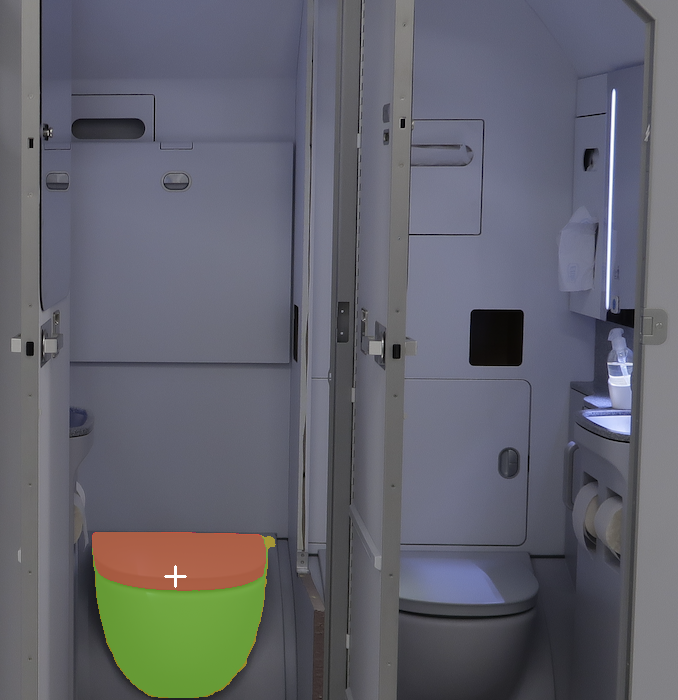}
        \end{subfigure}
    \end{subfigure}
    \caption{Prompting an image of an aircraft cabin monument to select regions of surface defects with \textbf{P.2} prompts (indicated by white cross marks): (a) cabin monument consisting of different slide-in shelves and a lavatory, (b) shows the sub-part / part / whole masks resulting from prompting the shelf and (c) the toilet.}
    \label{fig:cabinmonument}
\end{figure}

Initially, the cabin monument depicted in \cref{fig:cabinmonument} was used – a galley / lavatory complex. Addressing process step (3) (s. \cref{fig:FlowDigram_QA}), single point prompting \textbf{P.2} to select the regions of defects resulted in object-accurate masks even with respect to the three masks levels sub-part / part / whole (s. \cref{fig:cabinmonument}). The semantic-less masks generated by the SAM capture the components at a level of accuracy sufficient for the considered use case.

\begin{figure}[ht]
    \centering
    \begin{subfigure}[t]{\linewidth}
        \centering
        \begin{subfigure}[t]{0.33\linewidth}
            \caption{Dent}
            \includegraphics[width=\linewidth]{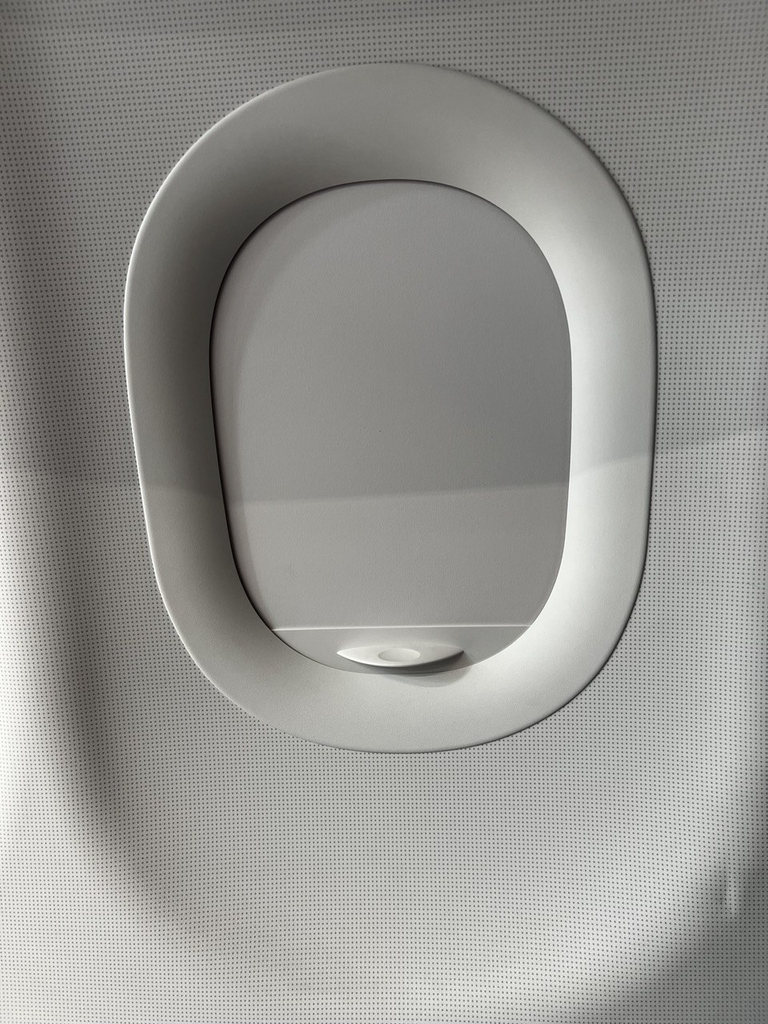}      
            \includegraphics[width=\linewidth]{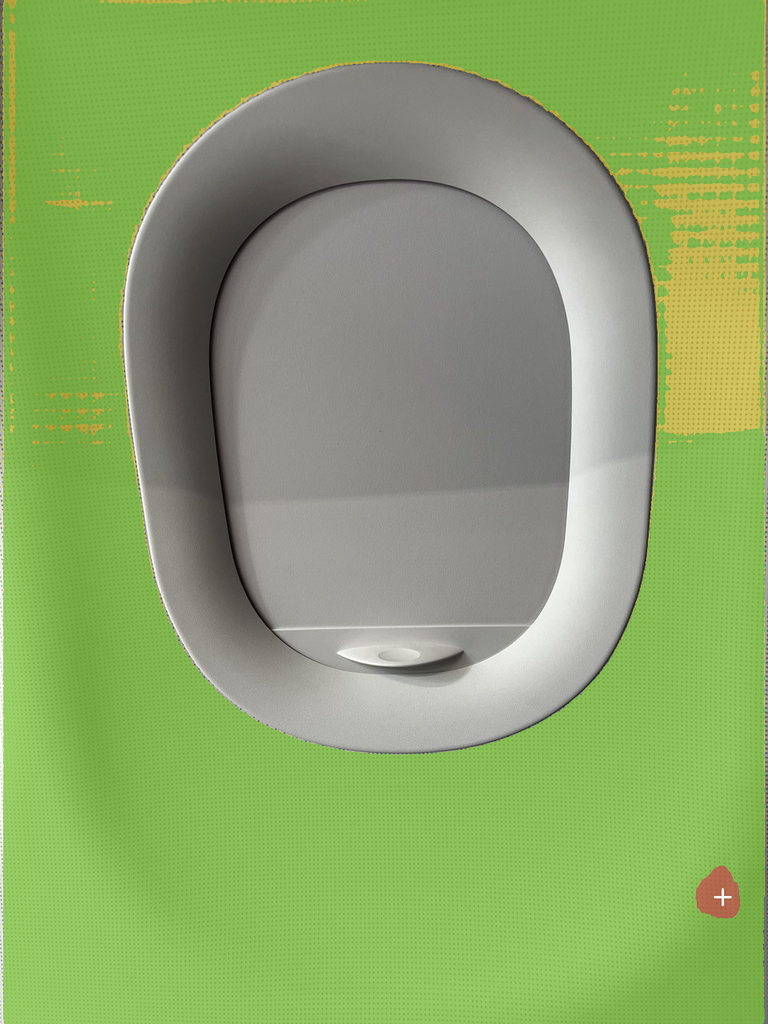}
        \end{subfigure}
        \begin{subfigure}[t]{0.33\linewidth}
            \caption{Dirt / Contaminants}
            \includegraphics[width=\linewidth]{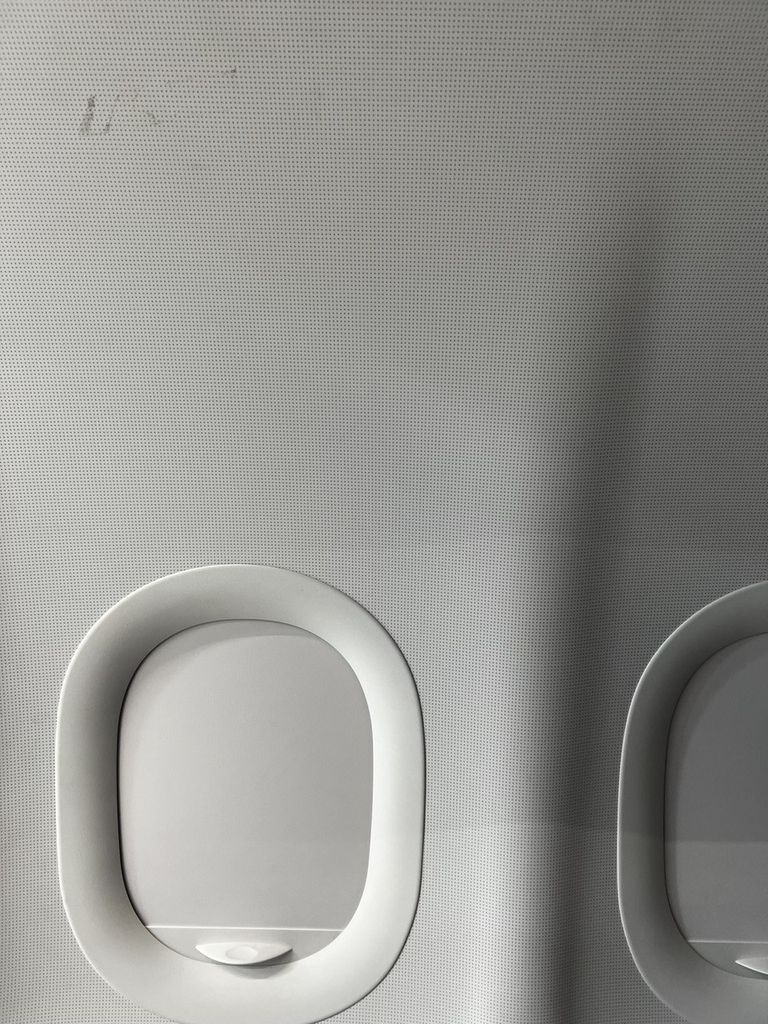}      
            \includegraphics[width=\linewidth]{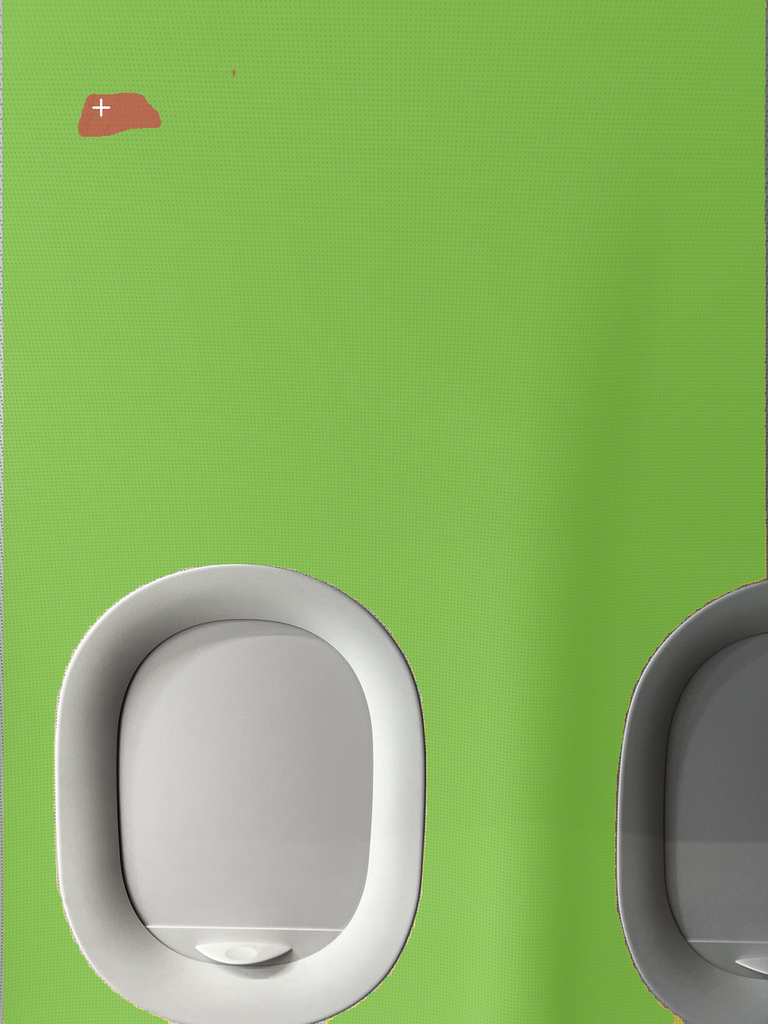}
        \end{subfigure}
        \begin{subfigure}[t]{0.33\linewidth}
            \caption{Indentations}
            \includegraphics[width=\linewidth]{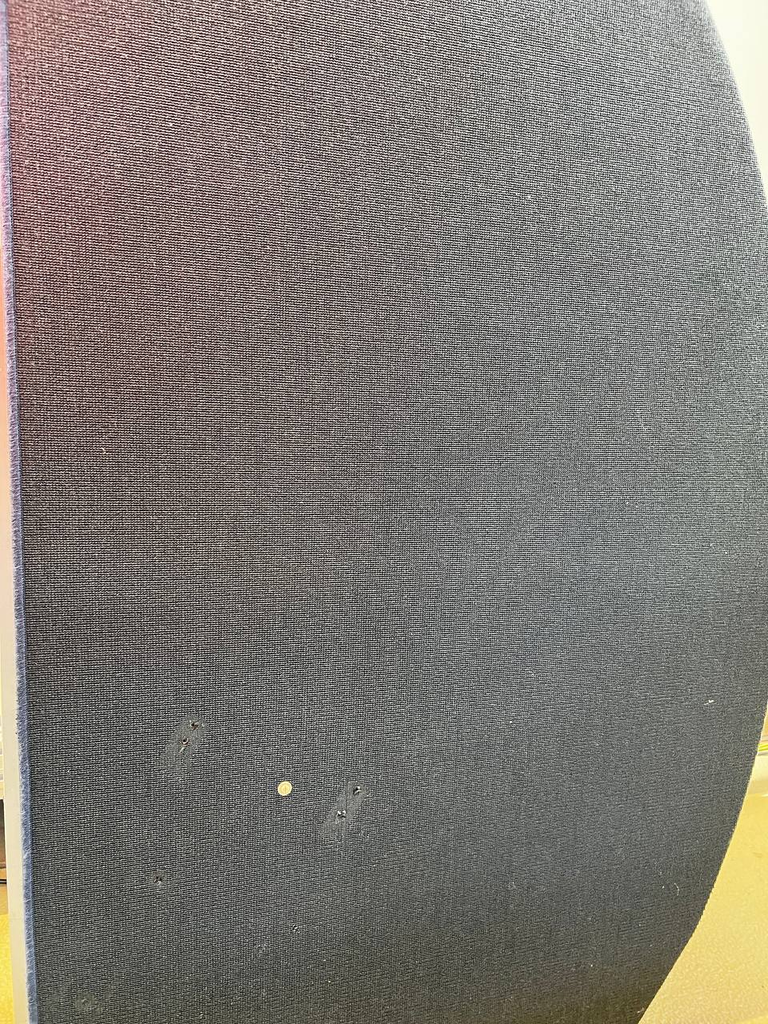}      
            \begin{subfigure}[t]{\linewidth}
                \includegraphics[width=0.329\linewidth,trim={0 0 2pt 0},clip]{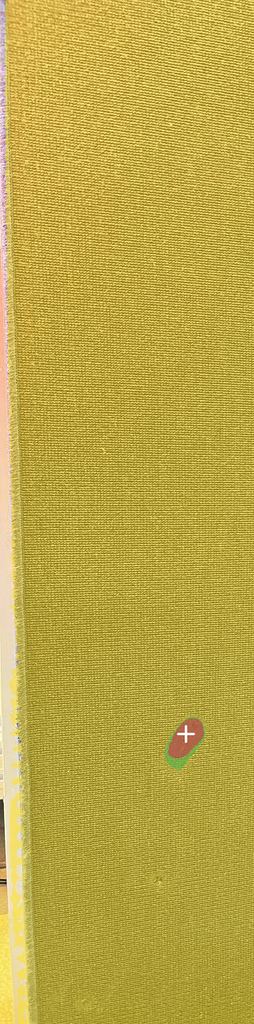}
                \includegraphics[width=0.658\linewidth,trim={2pt 0 0 0},clip]{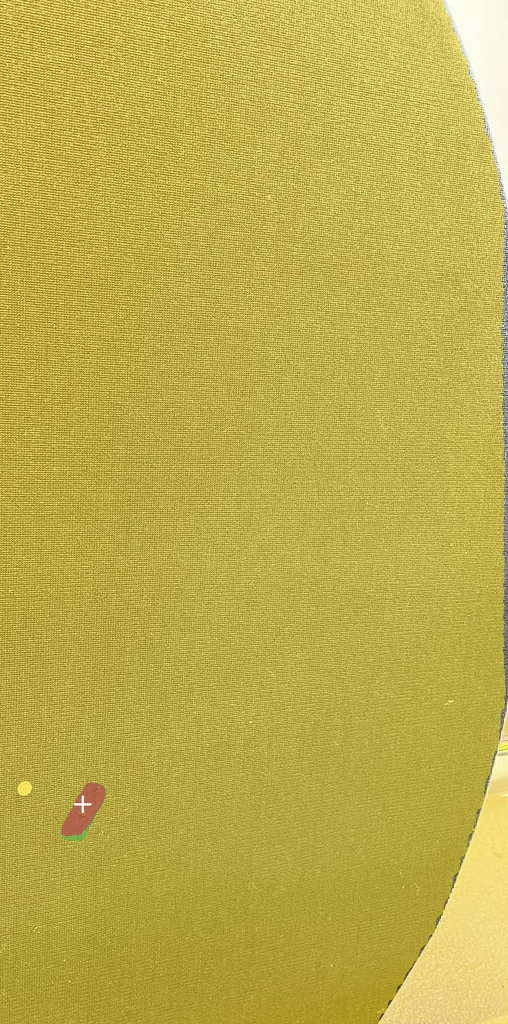}
            \end{subfigure}
        \end{subfigure}
    \end{subfigure}
    \caption{Prompting (\textbf{P.2}) different surface defects on cabin components: (a) dent on the surface of a Sidewall panel, (b) dirt / contamination, (c) two indentations on the outer fabric cover at the side of a slide-in shelf carrier (cropped into two separate images).}
    \label{fig:QA_defects}
\end{figure}

Further, outlining surface defects on individual components, which corresponds to process step (6) in \cref{fig:FlowDigram_QA}, was considered. Accurate marking of defects holds substantial importance as it contributes to quality documentation. In addition, accurate segmented masks of defects are an enabler for automated defect detection applications. For example, the theoretical accuracy of a supervised learning defect classification is related to the accuracy of the segmentation masks present in the training data. However, such precise manual segmentation as it is integrated into the considered quality control process is hardly economically feasible since complex defect geometries may be present depending on the defect class. 

As an example, three defect classes (dent, dirt / contamination, indentation) on two different component groups (Sidewall panel; carrier for slide-in shelves) were analyzed here (s. \cref{fig:QA_defects}). All the selected example images have in common that the defect region occupies only a relatively small image area of the photo and that they do not stand out significantly from the component background. 
Single point prompts \textbf{P.2} aimed at the defect region's center were utilized. In the case of the dent (a), the resulting „sub-part“ / primary mask accurately captures the defect geometry. However, the dirt is not accurately masked (b); major areas without dirt were also added to the segmentation mask. This insufficiently accurate mask may be considered irrelevant for the use case of quality documentation.
On the other hand, in the case of the indentation (c), the first two levels of masks cover the defect class accordingly.

For this use case, we did not use \textbf{P.3} prompts to refine the segmentation mask since we wanted to keep the segmentation process as simple as possible. With regard to the results of the fore- / background prompting in other use cases, it can be assumed that the quality of the mask can also be considerably improved with this prompting method.
In summary, simple point prompting, which uses a click input, contrasts with the time-consuming manual segmentation by inserting a polygon in the current application, possibly leading to a significant time advantage for accurate defect segmentation. This approach's robustness has yet to be proven by broader studies with more diverse defect classes and components. Furthermore, the usability must be evaluated since the system must reach a high acceptance by the inspection staff on the shop floor.

\section{Intralogistics \& Material Handling}\label{sec:use_cases_logistics}

Within the supply chain, intralogistics deals with the organization, execution, and optimization of the flow of resources within a specific location, such as a warehouse, distribution center, or manufacturing facility \cite{arnold_intralogistik_2006}, and supports the material flow. \emph{Storing}, \emph{transferring}, and \emph{handling} are the main functional elements of material flow \cite{lotter_montage_2006}, augmented by tasks as, e.g., \emph{quality assurance} and \emph{documentation} \cite{wehking_technisches_2020}. According to \cite{naumannLiteratureReviewComputer2023}, computer vision applications support logistics and material flow with applications in documentation, verification, assistance, and automation.

Due to a broad and variant-rich product spectrum and highly-complex assembly scenarios at the point of use, the logistics processes in the aircraft domain are predominantly executed manually. The processes for commissioning and transport are labor-intensive, repetitive, and physically straining for the logistics personnel involved and offer great potential for automation. Depending on the size and complexity as well as specific requirements for the transport, the material is either transported in standardized small load carriers (i.e., standardized transport boxes) or specialized load carriers unique to specific parts.

The remainder of this chapter will highlight different use cases in material handling along the value chain, starting with visual inspection of goods and load carriers for detection of waste, contamination, and parts left in returned load carriers in \cref{visual_detection_of_waste}. As a next step in the material flow, we highlight use-cases in parts commissioning and counting in \cref{picking_and_commissioning}, followed by applications in planning and execution of automated and autonomous material transfer in \cref{meterial_transfer}, as well as automated handling of load carriers and parts with bin-picking applications in \cref{robotic_bin_picking}.

\subsection{Visual Detection of Waste, Contamination, Damage, and Unassembled Parts}\label{visual_detection_of_waste}
Quality control, completeness, and content checks play an important role in logistics, as they do for manufacturing (s. \cref{sec:use_cases_manufacturing}) and MRO (s. \cref{sec:use_cases_mro}). The identification and documentation of defects or damages \cite{yang_detecting_2020, malyshev_artificial_2021}, tampering \cite{noceti_multi-camera_2018} and missing parts at early stages in the supply chain is desirable such that countermeasures can be taken. 
Quality assurance processes currently take place in a mainly manual fashion and could be (partially) automated or used for assistance systems by means of powerful segmentation using automatic grid prompting (\textbf{P.1}) and textual (\textbf{P.5}) prompts.

\begin{figure}[t]
    \centering
    \begin{subfigure}[t]{\linewidth}
        \centering
        \begin{subfigure}[t]{0.495\linewidth}
            \caption{Parts counting and bin picking}
            \includegraphics[width=\linewidth]{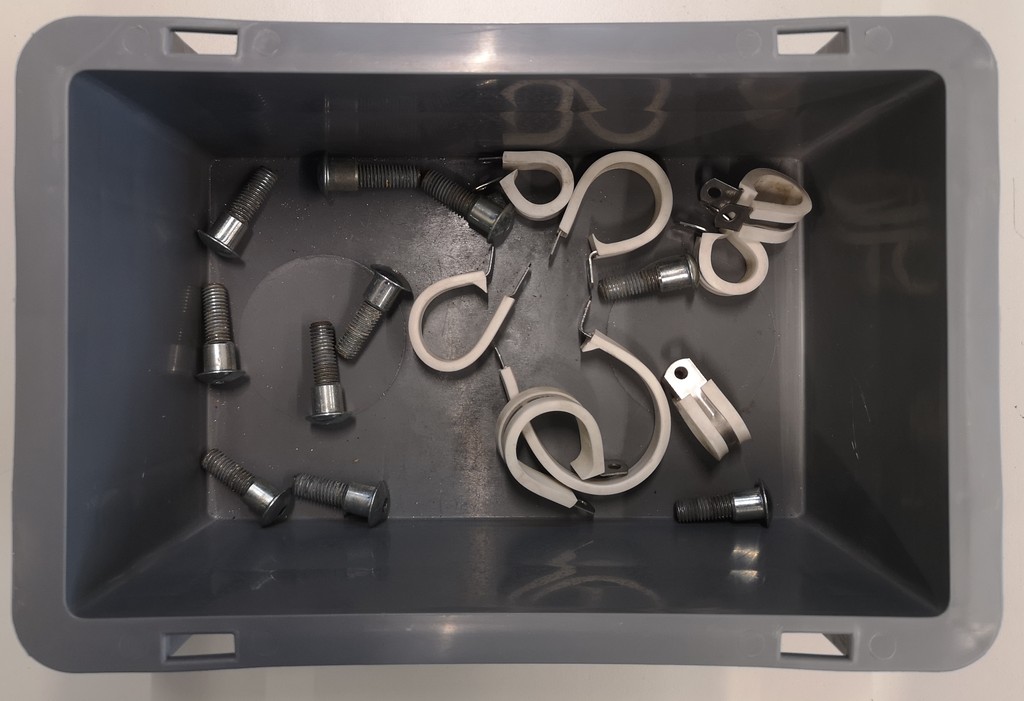}      
            \includegraphics[width=\linewidth]{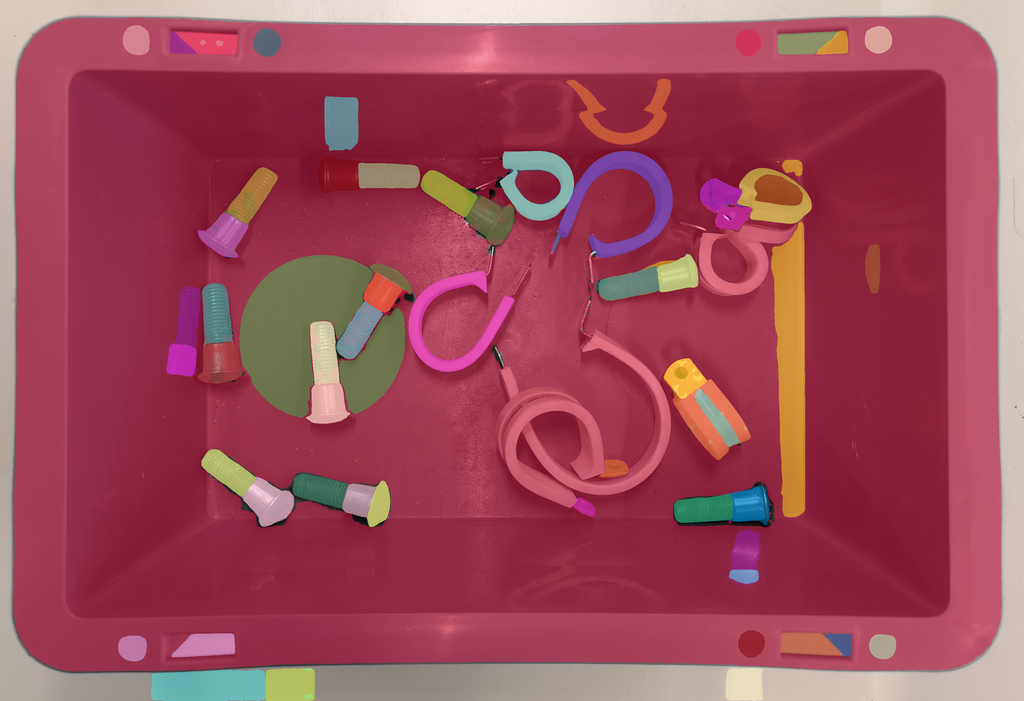}
        \end{subfigure}
        \begin{subfigure}[t]{0.495\linewidth}
            \caption{Object detection in small load carriers}
            \includegraphics[width=\linewidth]{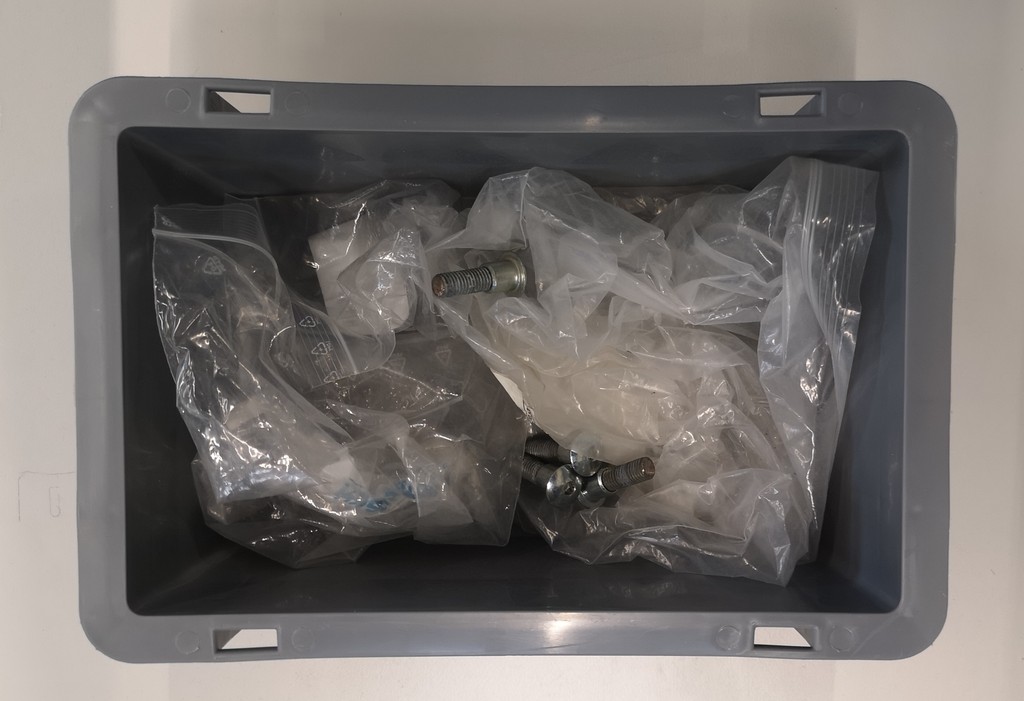}  
            \includegraphics[width=\linewidth]{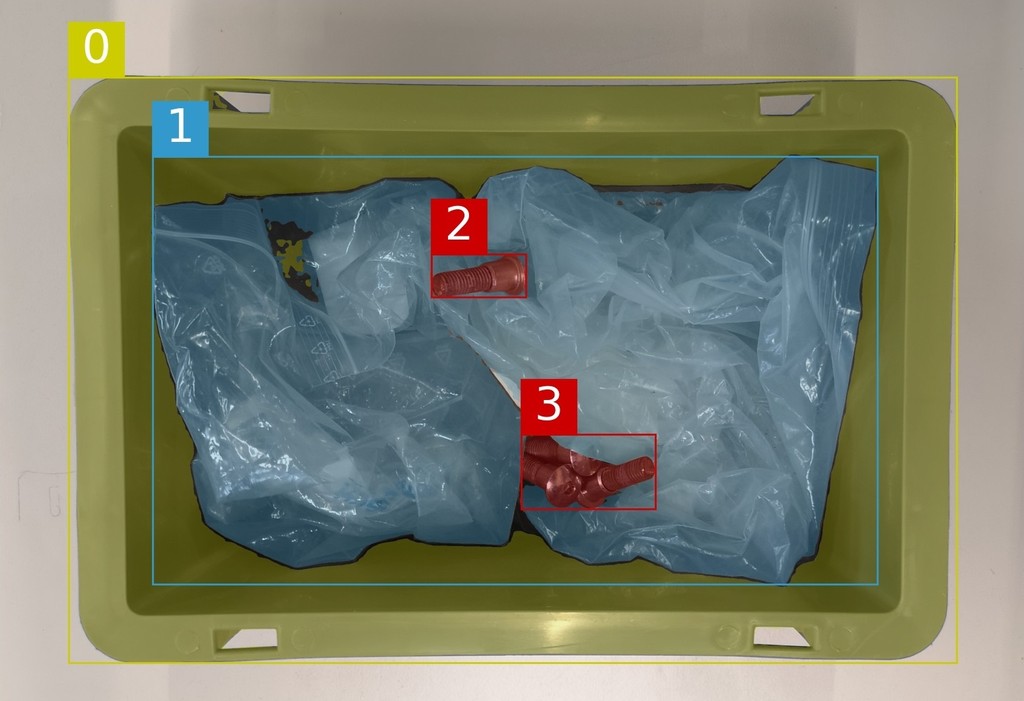}
        \end{subfigure}
    \end{subfigure}
    \caption{Segmentation allows for parts counting of unknown objects and helps to enable robotic bin picking of preliminary unknown objects (a), using automatic (\textbf{P.1}) mask generation. The application of the GroundedSAM (\textbf{P.5}) for the detection of packaging waste and remaining components is shown in (b). We prompted \textit{plastic. box. metal. plastic wrap. load carrier. machine part. fastener} resulting in masks for the load carrier, plastic wrapping, and fasteners.
    }
    \label{fig:logistics_use_cases}
\end{figure}

A typical application in intralogistic processes is the inspection of load carriers for trash, contamination, damage, and unassembled components -- often done manually.
Automated, early inspection for trash and damage offers the potential to streamline the process. 
Since the type and appearance of trash or damage is unknown in advance - a typical use case for anomaly detection, in conjunction with the SAM as discussed in \cite{cao_segment_2023}.
Experimentally, we used GroundedSAM\cite{noauthor_grounded-segment-anything_2023} for early visual identification of trash and unassembled parts using textual prompts, e.g., "plastic wrapping", "fastener" or "clamp".
Waste and components could then be iteratively removed with robots and fed to the respective downstream processes. 
Repeated segmentation, inspection, and removal also allow finding parts occluded or hidden below other materials. The right column in \cref{fig:logistics_use_cases} visualizes the segmentation results using the GroundedSAM. It can be observed that the model is capable of detecting unassembled parts laying on top of leftover plastic wrapping even though the fasteners are slightly occluded.

\begin{figure}[t]
    \centering
    \begin{subfigure}[t]{0.33\linewidth}
        \caption{Input data}
        \includegraphics[width=\linewidth]{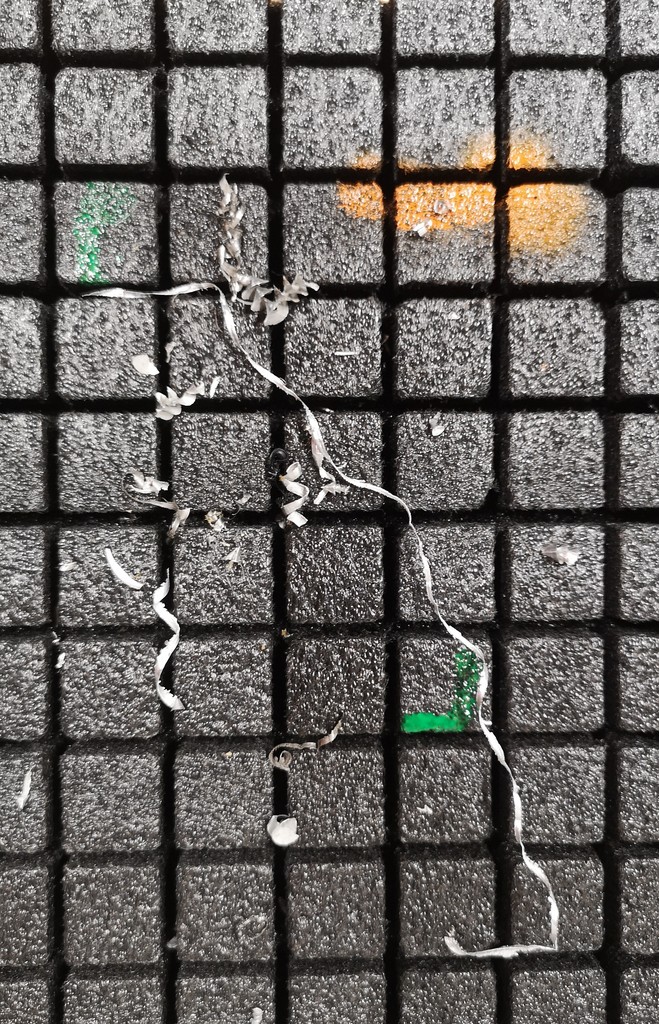}
    \end{subfigure}
    \begin{subfigure}[t]{0.33\linewidth}
        \caption{Default parameters}
        \includegraphics[width=\linewidth]{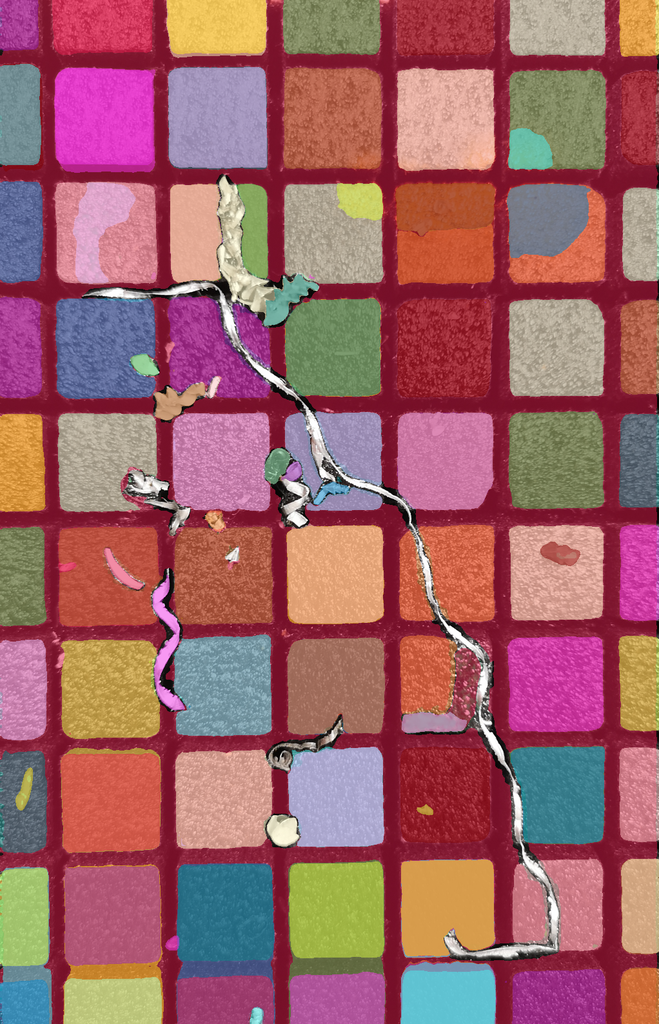}
    \end{subfigure}
    \begin{subfigure}[t]{0.33\linewidth}
        \caption{Optimized parameters}
        \includegraphics[width=\linewidth]{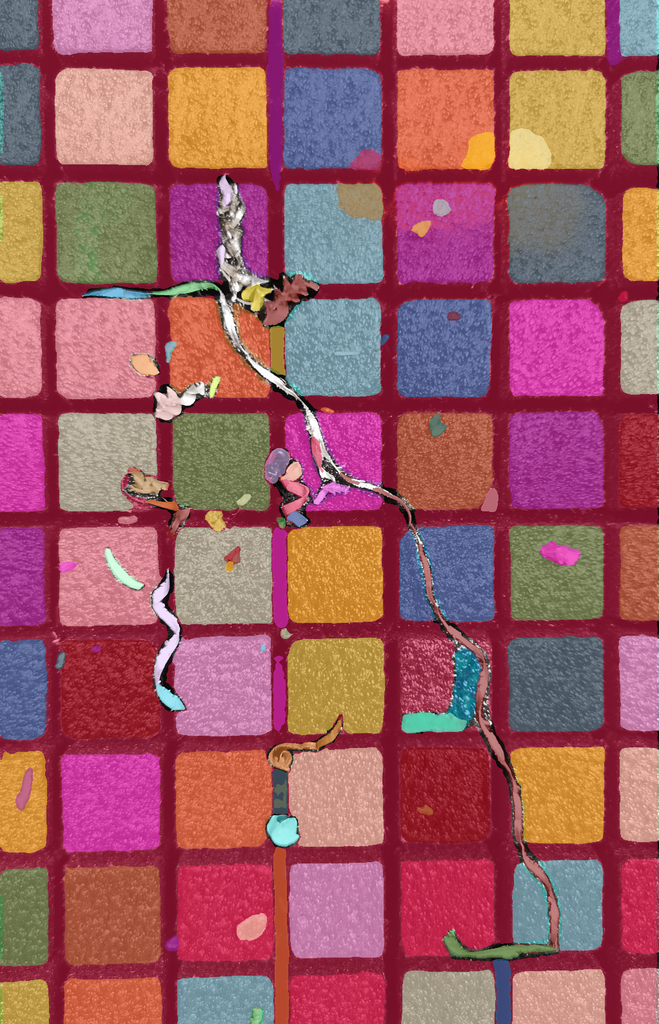}
    \end{subfigure}
    \caption{Foam inlay polluted with metal chips and color splashes (a). (b) and (c) showing overlays of the automatically generated segmentation masks (\textbf{P.1}), using default parameters (b) and the parameter set ($64$ points per side, $0.7$ pIoU, $0.2$ NMS threshold, $0.5$ stability score threshold, $250$ minimum mask region area) for (c).}
    \label{fig:FoD_foam_inlay}
\end{figure}

Another potential application is the visual inspection of foam inserts, which often get damaged or contaminated with metal shavings, paint, and sealing material. Contamination has to be identified and is tolerable up to a certain point.
In \cref{fig:FoD_foam_inlay}, the segmentation results of a polluted foam insert are shown. Using automatic prompting \textbf{P.1}, we observe differences in performance between the default parameters (\cref{fig:FoD_foam_inlay}.b) and a tuned set of parameters (\cref{fig:FoD_foam_inlay}.c). Both cases show accurate masks of the foam cubes, which can be used to detect foreign objects and contamination by comparing them with the expected outcome of a regular grid of square masks. 
Regarding the direct segmentation of chips, we observe the automatic segmentation with default parameters to show lower performance than the tuned parameter set. The variance of geometries, and the internal variance of features within the geometries (chips), require the use of lower confidence scores to be accounted for in the final set of masks. Nonetheless, most of the contamination is visible in both images or may at least be extrapolated from the cubes masks using domain knowledge.

\subsection{Picking and Commissioning}\label{picking_and_commissioning}
The recognition and identification of objects and load carriers for track \& trace applications, completeness checks, and documentation purposes are one of the most frequently performed computer vision-based tasks in material handling. Commonly found in inwards good inspection and documentation, and commissioning for material supply to processes further down the value chain. As already described in \cref{sec:use_cases_manufacturing}, visual markers, labels, or engravings are often used for this purpose, and sufficiently robust and proven systems exist for further processing.
To enable such tasks without markers as they occur for order picking or packaging processes as well as for (partially) automated material handling, a powerful segmentation of the objects in the cluttered scene can be an important pre-processing step. A following component recognition is then much more robust while being less complex and more task-specific. A big advantage is also the possible reduction of necessary training data since a renunciation of contextuality can be achieved. Currently, such tasks using computer vision and artificial intelligence are often enabled through extensive synthetic training data that aims to include all conceivable context \cite{schoepflinSmartMaterialDelivery2021, schoepflinSyntheticTrainingData2021, liu_towards_2021}.

Assortment boxes are used to commission small standardized parts that are required in high quantities, such as bolts, rivets, or clamps. Due to process requirements that arise from the high variability in the equipment of aircraft due to customer-specific demands, these assortment boxes contain more material than is necessary for a single aircraft. During the return, these boxes are not fully emptied, and the material has to be restocked according to the fill levels. Due to the varying quantity of returned parts, the number of parts to be refilled is not constant and has to be determined individually and counted accordingly. Particularly for small components and for large quantities, manual counting using a scale is time-consuming and error-prone. This can be remedied by counting machines, which usually separate the components and convey them along a sensor. Visual recognition through image segmentation is also a frequently used method for component counting.
The SAM's outstanding zero-shot segmentation capabilities shows promising results in the depicted domain (s. \cref{fig:logistics_use_cases}). In particular, the segmentation of composite parts, e.g., a clamp consisting of a metal clamp, rubber and screw, can be performed well with the SAM. Initial tests show that even overlapping and interlocked components are masked accordingly.

\subsection{Environment Analysis for Autonomous Materials Delivery}\label{meterial_transfer}
Environment analysis is another group of applications, of computer vision in logistics. Applications include free storage capacity utilization analysis, load handling device availability \cite{ozgur_comparing_2016}, precise localization of materials to be transported \cite{dorr_fully-automated_2020, mohamed_detection_2020}, and as a general means of perception for autonomous transportation vehicles and handling devices. Early-stage segmentation using the SAM can enhance such applications and boost developments with zero-shot segmentation capabilities for environment analysis and scene context understanding. For the delivery of material and tools by Automated Guided Vehicles (AGV)s or Autonomous Mobile Robots (AMR)s, as well as for robotic material handling of components in trolleys or boxes, a good knowledge of the environment is of great importance. Typically, such vehicles navigate by means of visual and depth sensor technology, using reconstruction and comparison of objects and map data with reference data. In addition, they often identify special landmarks, e.g., QR tags or characteristic stationary objects.
The identification, classification, and partly even localization of objects in the robot’s field of view is often performed using neural networks. These are trained using specialized, extensive training data for the application area. Fast pre-segmentation with the SAM can be particularly helpful for rapidly changing environments and those in which a high degree of object variability is encountered. The isolated object segments can then be quickly classified and efficiently localized individually. Segmentation and subsequent classification can thus be used to understand the scene piece by piece and eliminate the need to create extensive training data of all conceivable scenes since the segments can be processed in isolation from each other.

\begin{figure}[t]
    \centering
    \begin{minipage}{\linewidth}
        \centering
        \begin{subfigure}[t]{0.33\linewidth}
            \caption{Stitched input data}
            \includegraphics[width=\linewidth]{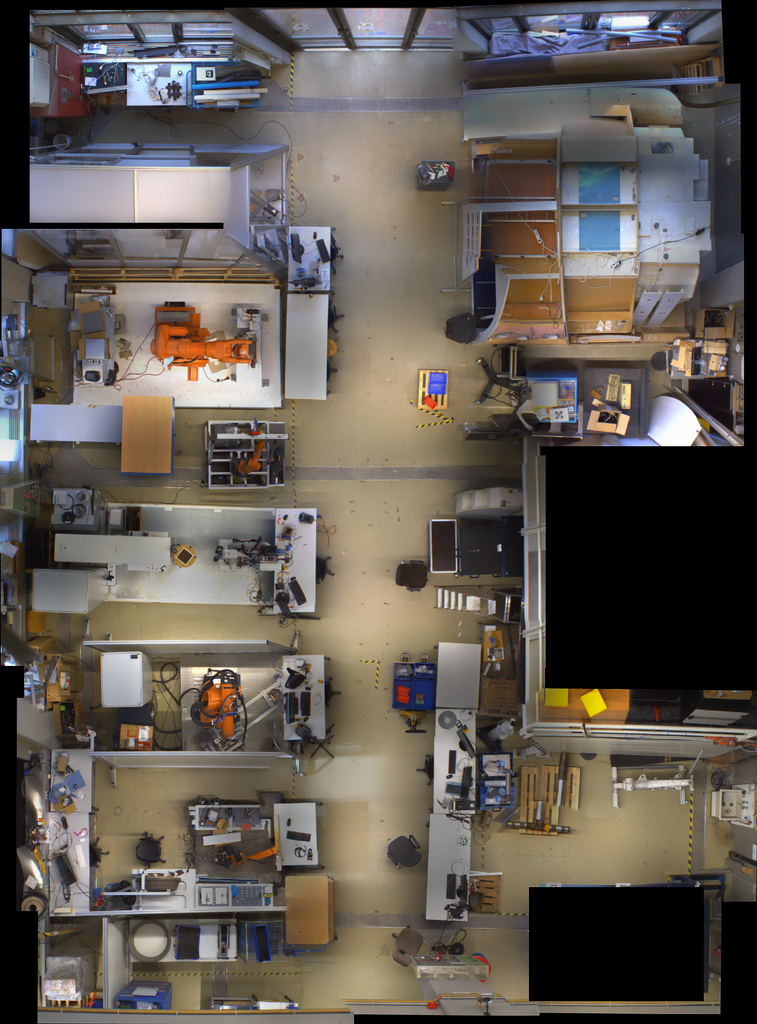}
        \end{subfigure}
        \begin{subfigure}[t]{0.33\linewidth}
            \caption{Automatic segmentation}
            \includegraphics[width=\linewidth]{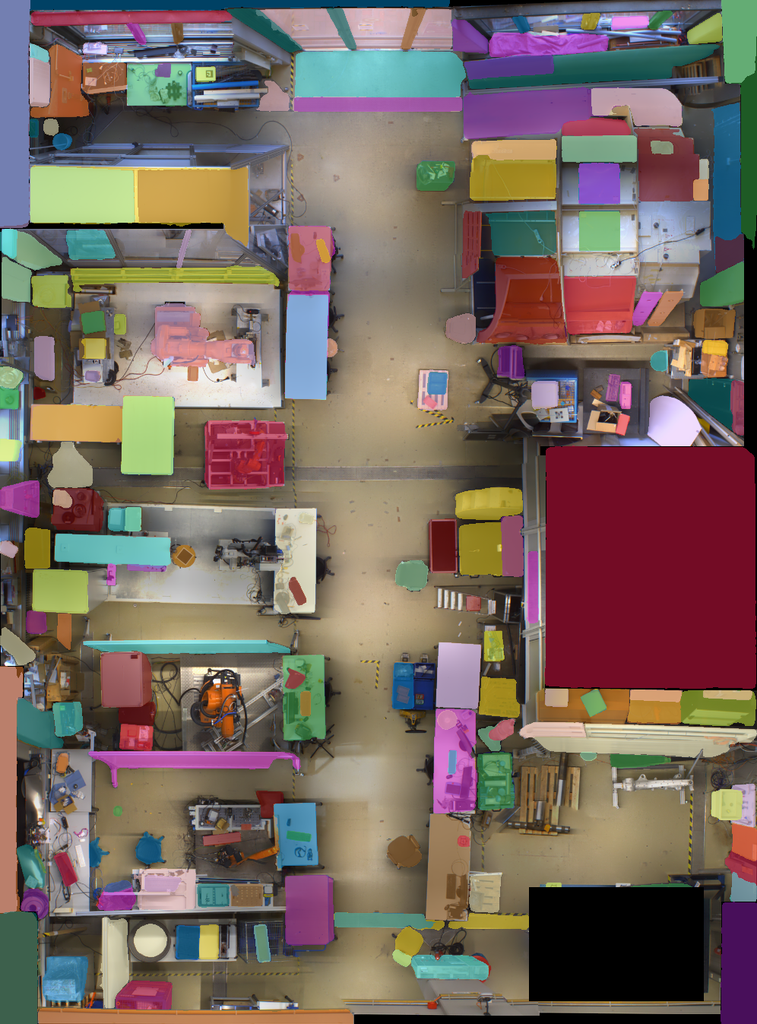}
        \end{subfigure}
        \begin{subfigure}[t]{0.33\linewidth}
            \caption{Floor segmentation example}
            \includegraphics[width=\linewidth]{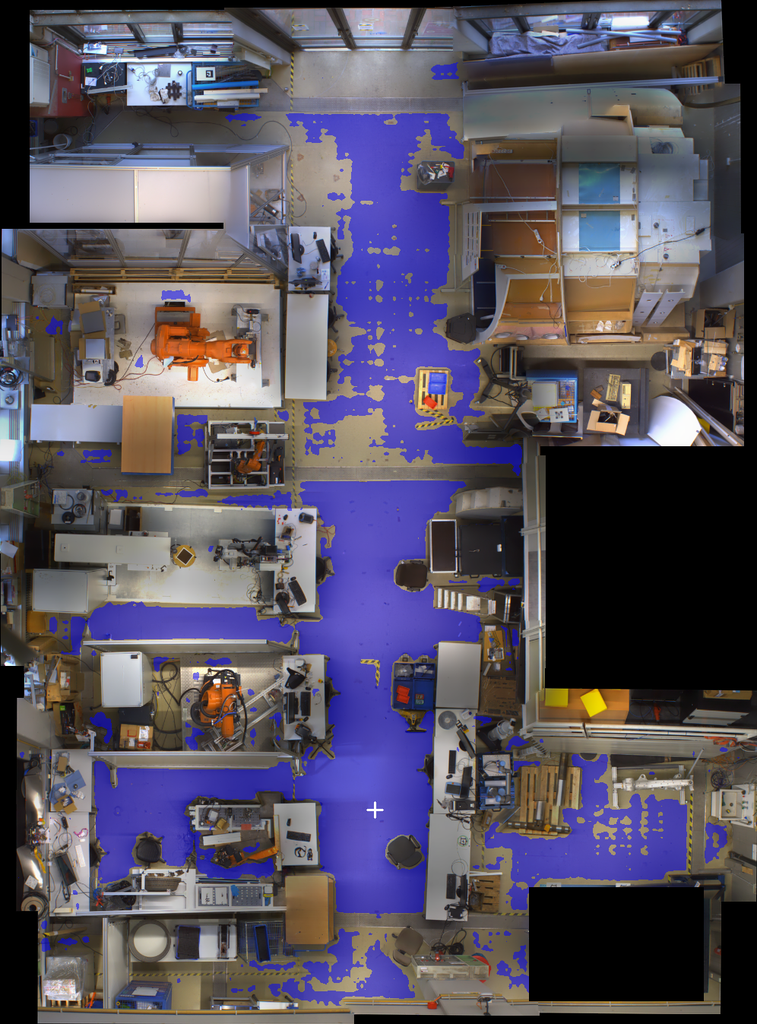}
        \end{subfigure}
    \caption{Applying automatic segmentation (\textbf{P.1}) and single location foreground prompt segmentation (\textbf{P.2}) to a bird's-eye view of a workshop. The input data (a) is stitched from 21 images of a ceiling-mounted 8$\times$3 camera array with a size of $996 \times 1347px$. (b) shows an overlay of the generated masks of an automatic segmentation (\textbf{P.1}, $64$ points per side, $0.9$ pIoU, $0.5$ NMS threshold, $250$ minimum mask region area) onto the source image. (c) shows an overlay of the largest segmentation mask, of a single location foreground prompt (\textbf{P.2}). The empty, black region in the lower right corner is redacted for legal purposes, after prompting and mask overlay. The blank region in the middle right is omitted from stitching due to its geometry and to reduce stitching complexity, thus present in the source data when prompting.}
    \label{fig:ceiling}
    \end{minipage}
\end{figure}

\begin{figure}[t]
    {
        \centering
        \renewcommand{\arraystretch}{5}
        \setlength{\tabcolsep}{0.006666\linewidth}
        
        \noindent\makebox[\textwidth]{
        \begin{tabular*}{\linewidth}{lcr}
            \begin{subfigure}[t]{0.32\linewidth}
                \caption{Input data}
                \includegraphics[width=\linewidth]{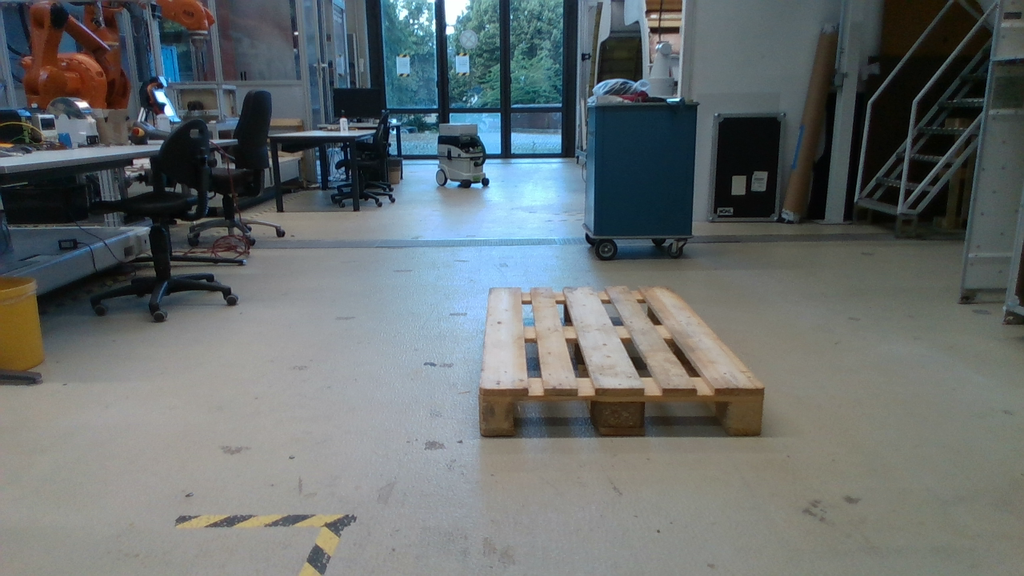}
            \end{subfigure}&
            \begin{subfigure}[t]{0.32\linewidth}
                \caption{EPAL segmentation}
                \includegraphics[width=\linewidth]{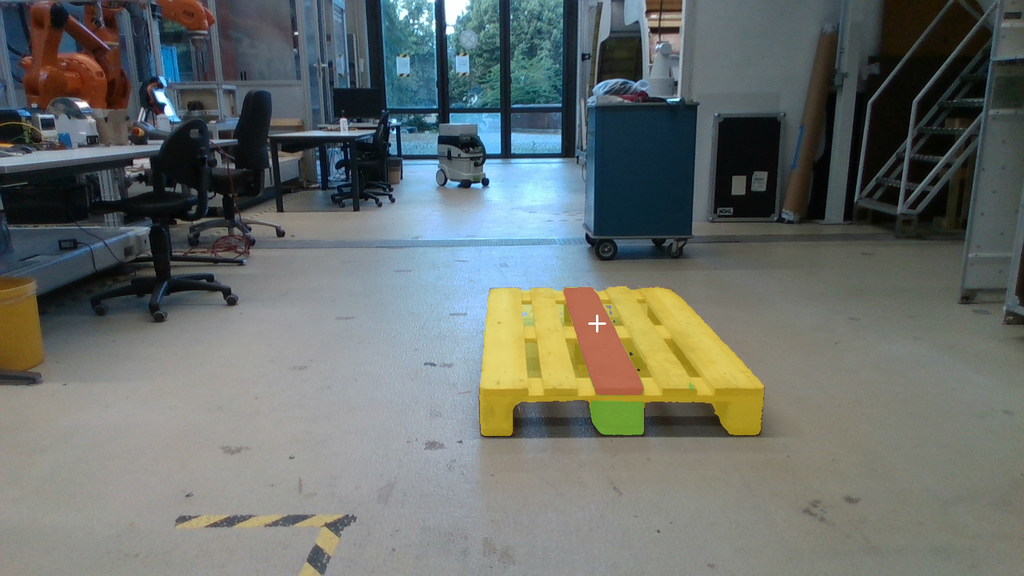}
            \end{subfigure}&
            \begin{subfigure}[t]{0.32\linewidth}
                \caption{Floor segmentation}
                \includegraphics[width=\linewidth]{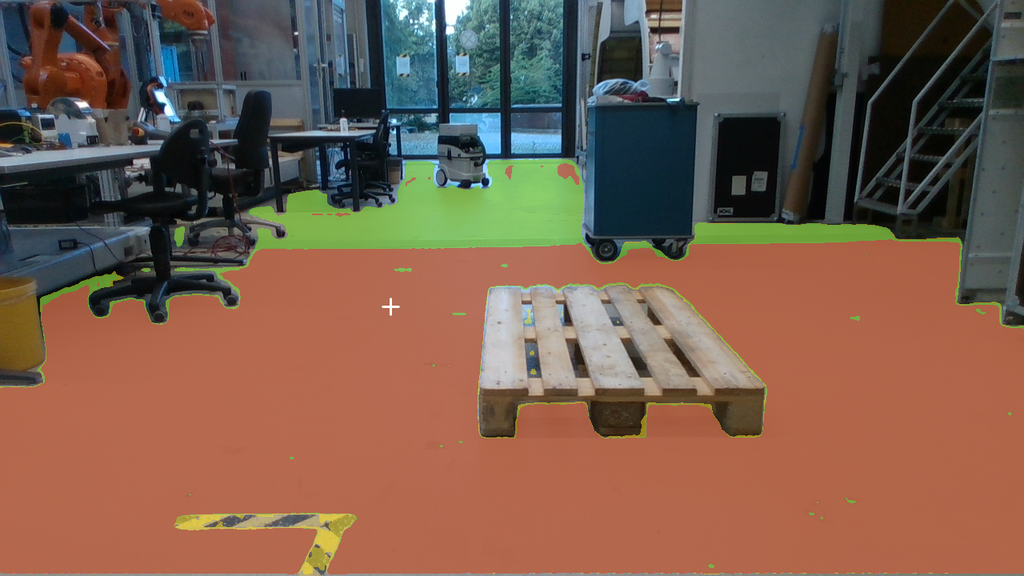}
            \end{subfigure}\\
            &
            \begin{subfigure}{0.32\linewidth}
                \includegraphics[width=\linewidth]{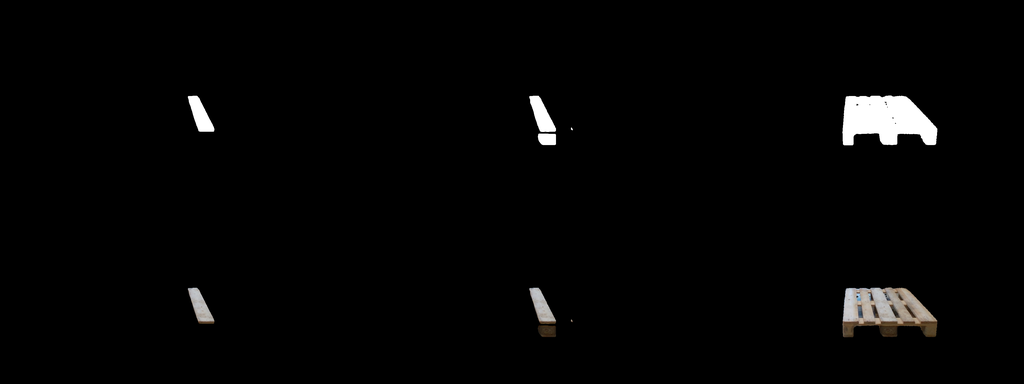}
            \end{subfigure}&
            \begin{subfigure}{0.32\linewidth}
                \includegraphics[width=\linewidth]{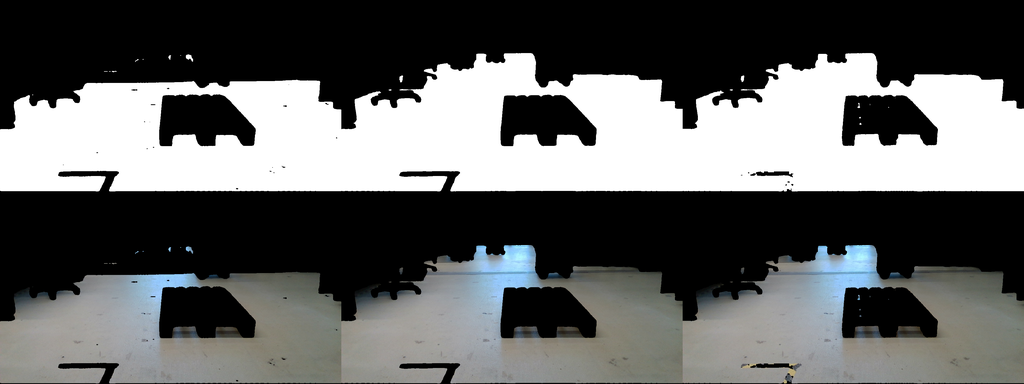}
            \end{subfigure}\\
            \begin{subfigure}{0.32\linewidth}
                \includegraphics[width=\linewidth]{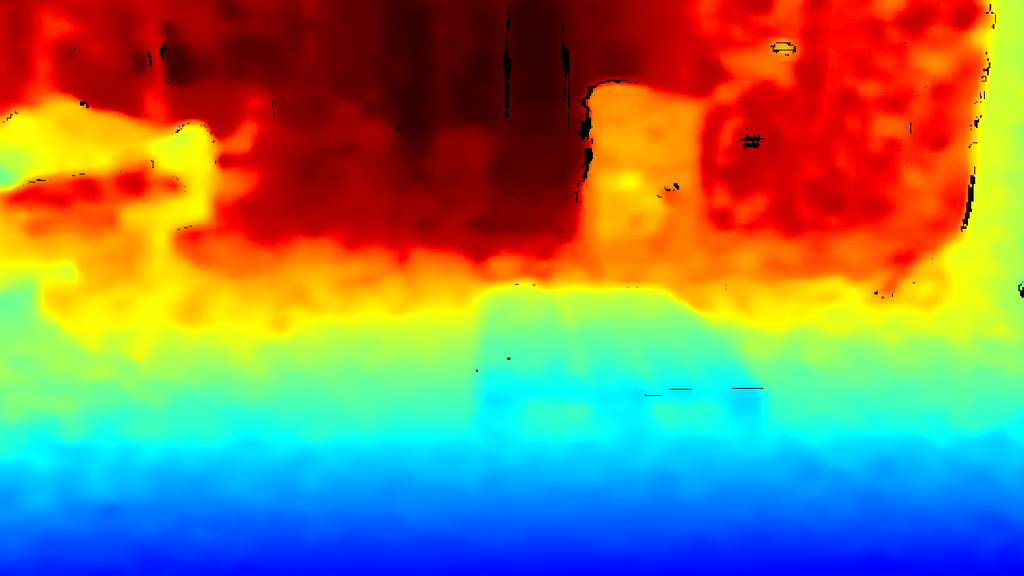}
            \end{subfigure}&
            \begin{subfigure}{0.32\linewidth}
                \includegraphics[width=\linewidth,trim={0 10em 0 10em},clip]{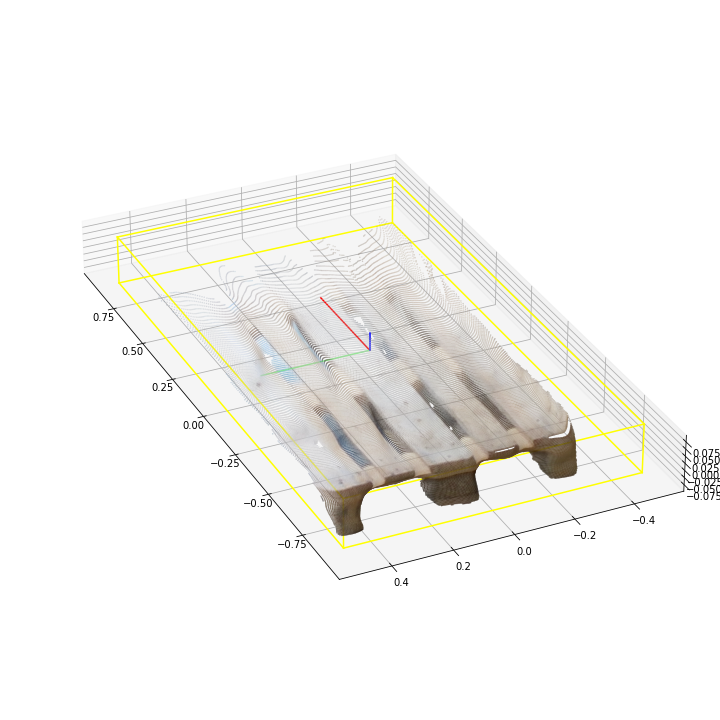}
            \end{subfigure}&
            \begin{subfigure}{0.32\linewidth}
                \includegraphics[width=\linewidth,trim={0 10em 0 10em},clip]{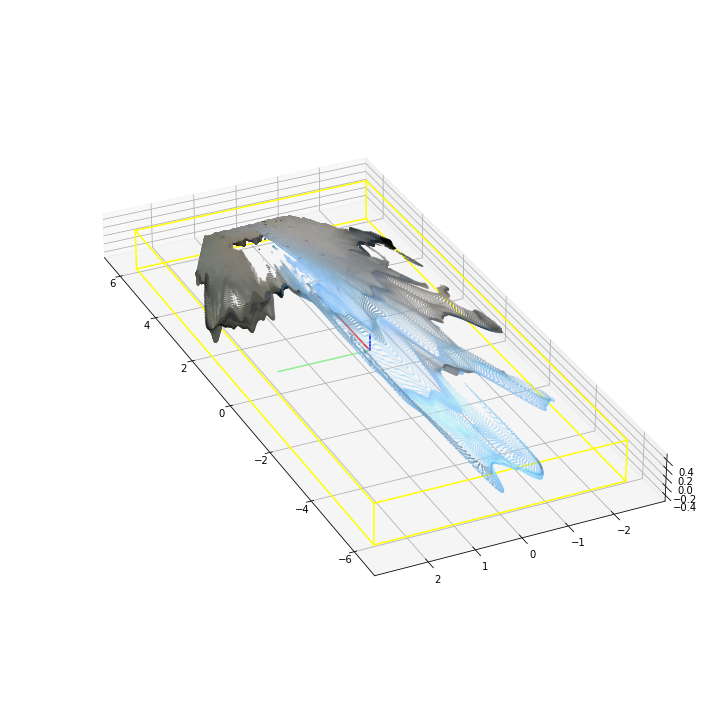}
            \end{subfigure}\\
        \end{tabular*}
        }
    }
    \caption{3D pointcloud segmentation from aligned depth and image data, captured with an \emph{Intel Realsense D435} at $\scsim 1m$ height pointing downwards at $\scsim 20\degree$. (a) shows the input image (top) and aligned depth data as 2D colored image (bottom). (b) and (c) show overlays of the input data with segmentations masks generated from single point prompting (\textbf{P.1}), followed vertically by all three generated segmentation masks and the result of applying the mask to the input image. At the bottom, (b) and (c) show the colorized point cloud, obtained by applying each largest segmentation mask to the depth data. The point clouds are transformed to align with the origins' axes and have the centroid of their oriented bounding box at the origin, for visualization purposes.}
    \label{fig:3d_logistics}
\end{figure}

Using four different source datasets, in this section, we want to highlight use cases in the realm of autonomous materials delivery, using the SAM on 2D- and combined 2D/3D-data. We illustrate possibilities for the application of the SAM to optimize processes from system setup to mission planning and use in path planning and navigation, as well as manipulation and handling of load carriers. Handling of individual parts and automated robotic grasping is considered in \ref{robotic_bin_picking}. In addition to the examples provided in this section, \cref{sec:3D_scans} will provide a more in-depth view on the use of the SAM for 3D-data.

Performing autonomous operations requires a frame of reference for all participants and systems (hardware and software), from planning to execution, usually referred to as a map. Without further specifying the type of map, for the use-case of intralogistics and material transport, it commonly includes at least information about the location and geometry of pick-up and drop-off locations for goods and a coarse map of the environment for mission planning and global path planning. Generating this information is a mostly manual task, from a (building-) design stage to the inclusion of as-built deviations later on in the process. While exploration and guided-exploration techniques can be used to facilitate AGVs' sensors to create maps, yielding mostly geometric occupancy-grid-based maps; without semantic annotations and within the constraints of accessibility due to vehicle kinematics, occlusion, and accuracy of the used -- usually -- SLAM algorithm. \Cref{fig:ceiling}, showing a bird's-eye view as a stitched 'panoramic' image of ceiling mounted camera, illustrates two uses in system setup for AGVs. We assume that qualitatively similar data may be obtained using UAVs/drones or projections of RGBD point clouds created by terrestrial scanning.
\Cref{fig:ceiling}.b shows the masks generated by an automatic segmentation of the input image \cref{fig:ceiling}.a. We observe a rather high probability for objects the size of medium-sized load carriers to the size of pallets and above to be segmented and see a strong case for their use in labeling and mapping the location and orientation of stations, pick-up, and drop-off zones for AGVs. Assuming this process to be \emph{human-in-the-loop}, an initial segmentation mask and location estimation can reduce the time needed to create the final dataset. \Cref{fig:ceiling}.c shows the segmentation with manual prompting at the shown and specified location. We observe the SAM to correctly classify disjoint regions of the floor as the same object. While the colors of the segmented region are similar, this is a clear advance over, e.g., flood-filling algorithms. This data can subsequently be used in initial mapping and system setup for geometric maps and used in mission and global path planning.

\Cref{fig:3d_logistics} shows image and depth data as would be captured by an AGV, with a sensor height of $\scsim 1m$ above ground, angled down at $\scsim 20\degree$. \Cref{fig:3d_logistics}.c shows the same principle as applied in \cref{fig:ceiling}.c, segmenting the floor as usable pathways for the AGV. Here we apply the obtained segmentation mask to the captured depth data as well, obtaining a segmented point cloud of the floor. In 2D, inverting the segmentation mask, or in 3D, subtracting the segmented point cloud from the input data yields obstacles the AGV has to avoid and consider in its (local) navigation. A method for automatically generating suitable locations for \textbf{P.2} prompting within the source image must be developed when embedding the SAM in a real-world application.

The SAM shows opportunities for assistance in manipulation and handling with automated and autonomous systems. \Cref{fig:3d_logistics}.b show the application to combine 2D and 3D data -- as outlined above -- but prompting (\textbf{P.2}) at the location of the pallet in the source data. The resulting point cloud can be used to determine the pose of the pallet for pick-up by an AGV. The segmentation of 2D data with subsequent application of the mask to the 3D data provides a high potential to extract use- and meaningful data with lower computational resources than required for inference directly on 3D datasets.

\begin{figure}[t]
    \centering
    \begin{subfigure}[t]{\linewidth}
        \centering
        \begin{subfigure}[b]{0.27\linewidth}
            \caption{Trolley}
            \includegraphics[width=\linewidth]{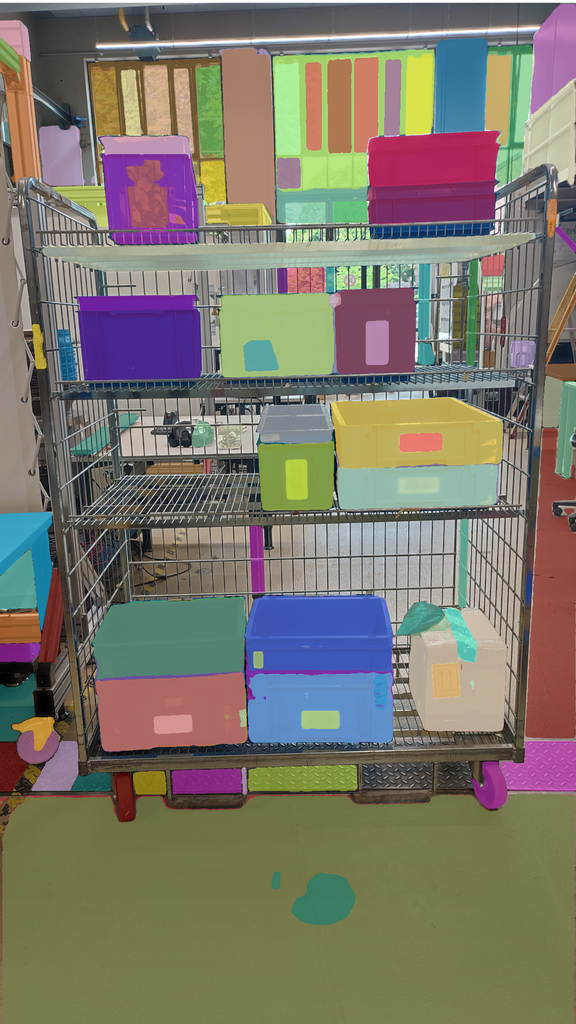}
        \end{subfigure}
        \begin{subfigure}[b]{0.71\linewidth}
            \caption{Pallet truck}
            \includegraphics[width=\linewidth]{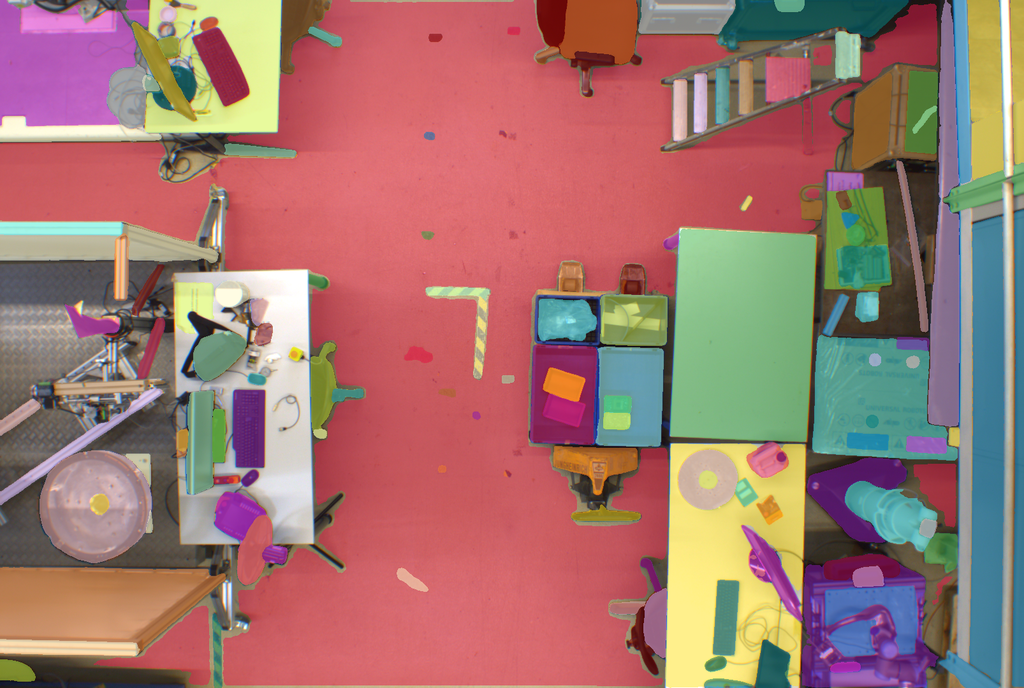}
        \end{subfigure}
    \end{subfigure}
    \caption{Both images (a), (b) show an overlay of the generated masks of an automatic segmentation (\textbf{P.1}) on the source image. (a) using default parameters (cmp. \cref{sec:sam}) with frontal view of a trolley, (b) using $64$ points per side, $0.9$ pIoU, $0.5$ NMS threshold, $250$ minimum mask region area with a bird's-eye view down onto the shop floor.}
    \label{fig:carrier_segmentation}
\end{figure}

Moving down the scale of load carriers -- from pallets to small load carriers -- \cref{fig:carrier_segmentation} shows overlays of automatically (\textbf{P.1}) generated segmentation masks of (\cref{fig:carrier_segmentation}.a) a material trolley with small load carriers and a cardboard box, and (\cref{fig:carrier_segmentation}.b, center) a pallet truck loaded with a pallet and a number of small load carriers stacked atop. For the trolley, we observe a $100\%$ segmentation rate of all present small load carriers, while for the pallet truck, no segmentation mask for the top left carrier is available, resulting in $87,5\%$ segmentation rate of the desired objects. The obtained segmentation masks can subsequently aid in the automated handling of the small load carriers, e.g., for automated unloading of the trolley by a robotics arm, by performing further analysis of the scene and calculating the carrier poses and potential obstacles and obstructions.
The obtained segmentation masks can further be used, not only for handling but for inspection and process guidance. Based on the trolleys' segmentation masks, remaining load carriers can be detected as enclosed thereby, and the state of loading or unloading be supervised and documented.

\subsection{Robotic Bin Picking of Unknown Objects}\label{robotic_bin_picking}
Computer vision can provide the necessary information to enable automated and flexible robotic handling tasks, up to the complexity of bin picking \cite{khalid_deep_2019, zhuang_semantic_2021, dong_ppr-netpoint-wise_2019, clausen_3d-computer_2014}. Visual object identification and the further identification of grasping positions are crucial for a fully operational robotic handling application. For subsequent picking and packing, computer vision applications allow packaging analysis and can determine suitable packaging strategies based on the volume and shape of the components and containers.

Due to the high degree of customization, complexity, and size of an aircraft, manufacturing is characterized by a wide variety of parts and variants on a multitude of scales. In addition to a certain part spectrum comprised of standardized parts that are also common in other industrial sectors, like standard fasteners, there is a great number of parts specific to the aircraft domain. That makes the introduction of widespread automation for part handling challenging. Globally speaking, training models for the full part spectrum seem uneconomical. For this domain, novel approaches for model-free and object-agnostic robotic handling seem more economical to allow for a future-proof and versatile system.

Robotic bin-picking applications, especially model-free approaches, rely commonly on the use of RGBD data for object detection and determination of grasping strategies. Segmentation of object instances here is a highly crucial step for general performance. In most solutions, this is the first processing step, and the quality directly influences following calculations up to grasping an object.
Focusing on the examples from \cref{fig:logistics_use_cases}, the mask robustness in (a) and (b) is probably sufficient for further model-free grasp candidate generation, especially when merging the RGB masks with the depth information if using an RGBD camera.

\section{Maintenance, Repair, and Overhaul (MRO)}\label{sec:use_cases_mro}

Aircraft MRO refers to a set of activities involved in ensuring airworthiness and operational readiness of an aircraft. It covers all activities dealing with the maintenance, repair, and overhaul, keeping the aircraft in safe and reliable condition over its entire operational life.

Maintenance involves scheduled inspections and preventive maintenance tasks, ensuring the aircraft meets all safety regulations and manufacturer’s recommendations. It includes routine checks, such as daily inspections, periodic inspections, and more extensive check-ups for which the aircraft is temporarily taken out of service.
Repair refers to activities that involve fixing any damage or malfunction identified during inspections. The activities can range from minor fixes or replacement of components to major structural repairs resulting from an accident.
The aircraft is extensively examined and re-manufactured during overhaul to restore it to a like-new condition. Components are disassembled, inspected for wear and tear, repaired, and exchanged in these checks. 

\cite{yasudaAircraftVisualInspection2022} present a literature review on visual inspection activities in aircraft MRO. Applications include, e.g., the detection of different defects on the aircraft fuselage and structure, such as automated dent detection \cite{bouarfa_towards_2020}, fuselage defect detection \cite{malekzadeh_aircraft_2017, ramalingam_visual_2019} and structure crack and damage detection \cite{li_yolov3-lite_2019, fotouhi_autonomous_2021}. There are also some works targeting specifically the aircraft engines as those need to be checked regularly for damaged fan blades and the combustion chamber using borescopes as such parts are not reachable without special equipment \cite{upadhyay_deep-learning-based_2023, shen_deep_2019}.

In the following \cref{sec:mro_visual_inspection}, we discuss two use cases related to visual inspection.
\cref{sec:mro_process_monitoring} discusses using the promptable segmentation model in process monitoring and control. Finally, \cref{sec:3D_scans} discusses the segmentation of 3D scans based on the projection to 2D images for the application in aircraft interior retrofitting.

\subsection{Visual Inspection}\label{sec:mro_visual_inspection}

\subsubsection{Borescopy}\label{sec:borescope_inspection}

Aircraft engine inspection plays a critical role in ensuring flight safety, as it aims to maintain the performance standards required to keep the engine airworthy. In addition, engine inspection is critical to maintaining the value of the asset. A key inspection is the borescope inspection (BI), as it allows visual inspection of the engine interior while the engine is assembled and on the wing.

For BI, a rigid or flexible tube with a high-resolution camera at one end is inserted through a small opening in the engine. The most frequently complained-about engine parts are blades and vanes from the compressor and turbine area. These blades are subjected to high loads resulting from extreme operating conditions, such as high centrifugal loads, high temperatures, high pressures, and vibrations.

The process of borescope inspection is characterized by time-consuming and subjective elements that make it a complex undertaking. The effectiveness of this inspection is highly dependent on the expertise and attention of the inspector throughout the process. To date, the BI is a largely manual task \cite{li_deep_2022}. The reporting of borescope images is a demanding task that has not yet been fully automated. AI-based image processing has a hard time in the field of BI because sufficiently powerful training data is not available.

\begin{figure}[t]
    \centering
    \begin{subfigure}[t]{\linewidth}
        \centering
        \begin{subfigure}[t]{0.195\linewidth}
            \caption{Corrosion}
            \includegraphics[width=\linewidth]{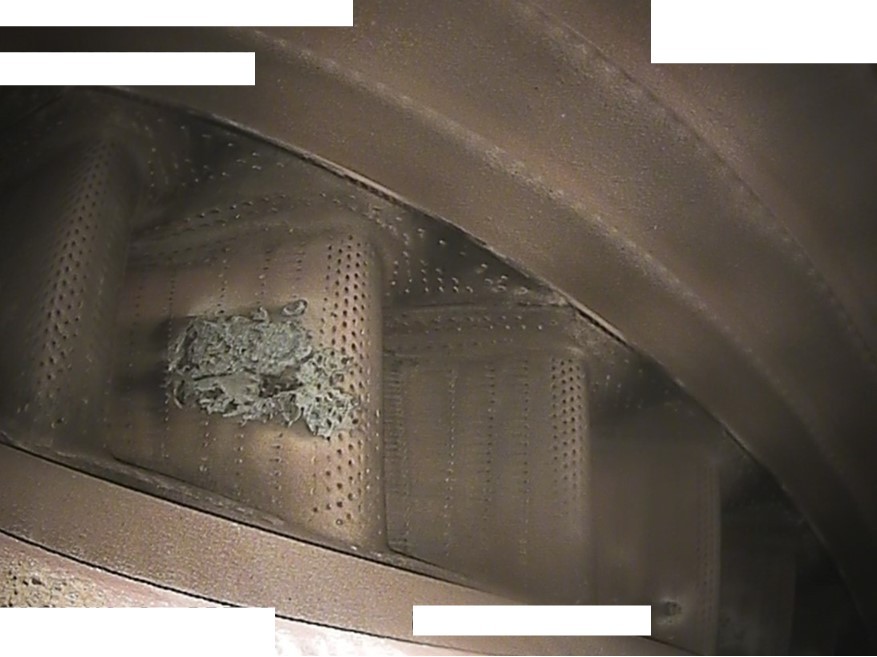}      
            \includegraphics[width=\linewidth]{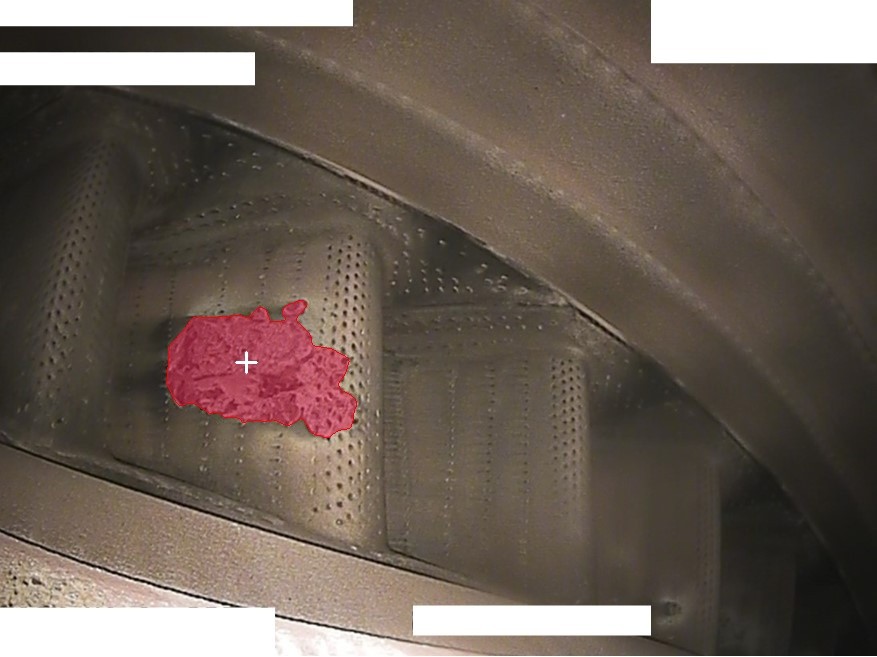}
        \end{subfigure}
        \begin{subfigure}[t]{0.195\linewidth}
            \caption{TBC missing}
            \includegraphics[width=\linewidth]{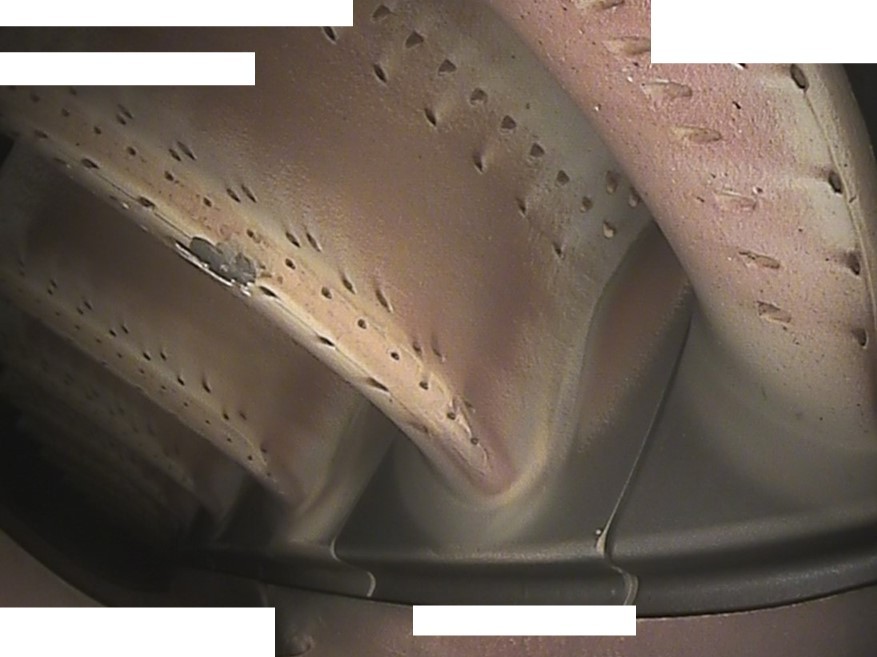}  
            \includegraphics[width=\linewidth]{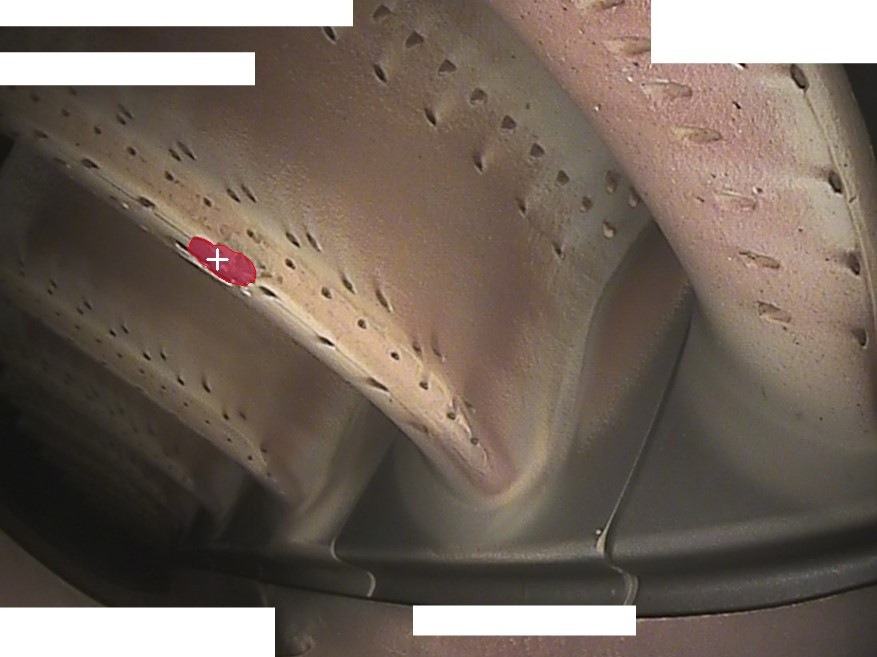}
        \end{subfigure}
        \begin{subfigure}[t]{0.195\linewidth}
            \caption{Crack}
            \includegraphics[width=\linewidth]{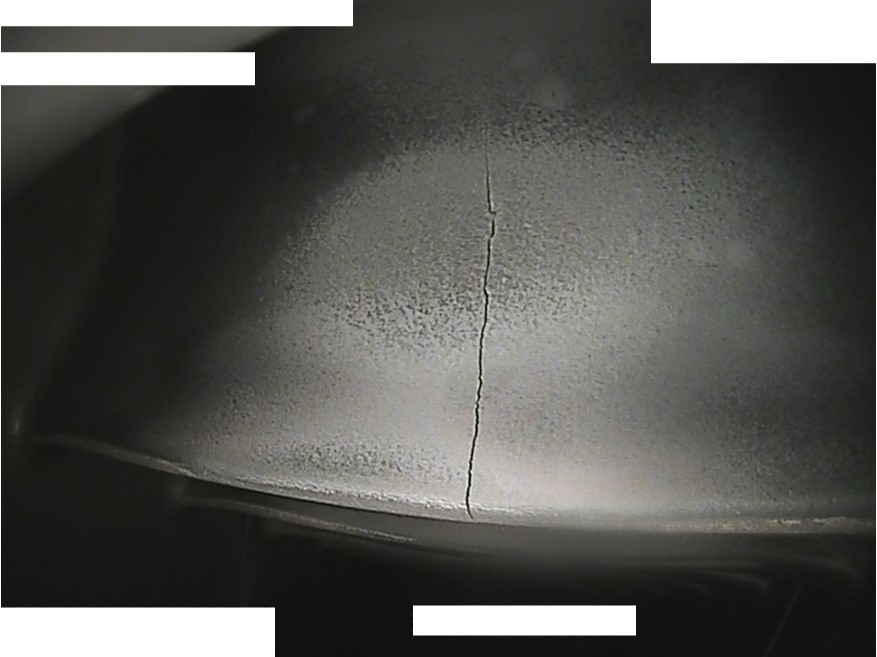}
            \includegraphics[width=\linewidth]{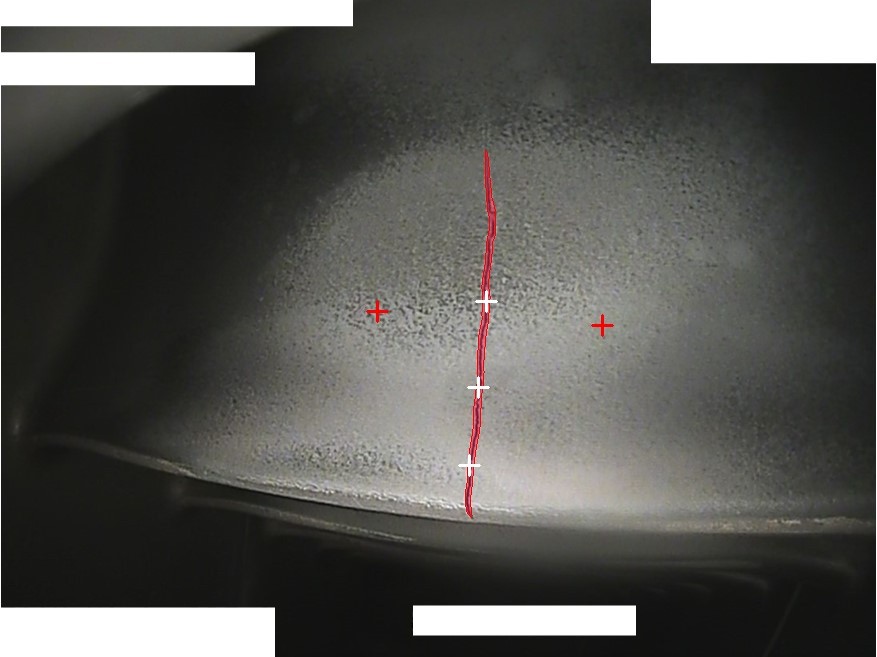}
        \end{subfigure}
        \begin{subfigure}[t]{0.195\linewidth}
            \caption{Dent}
            \includegraphics[width=\linewidth]{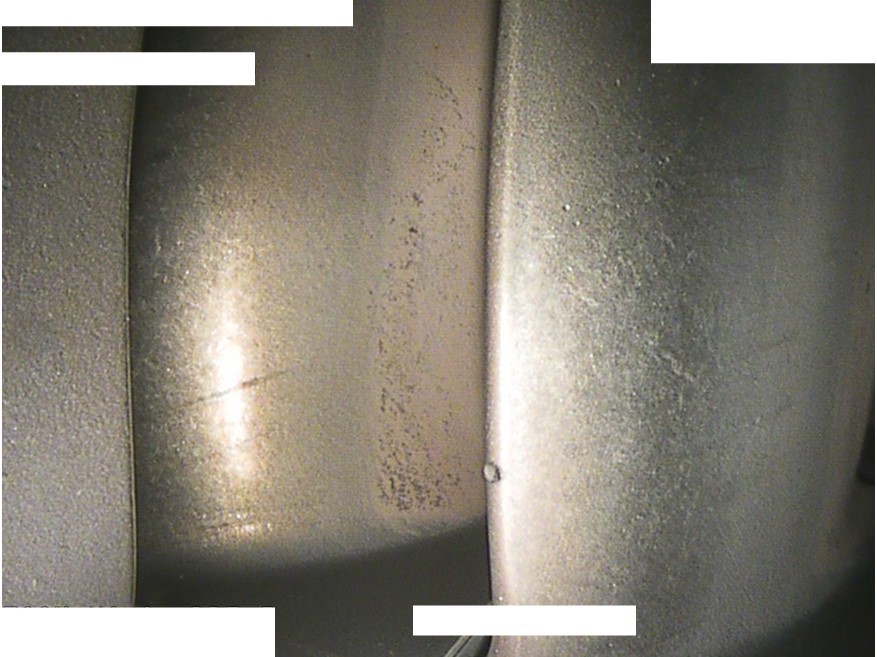}
            \includegraphics[width=\linewidth]{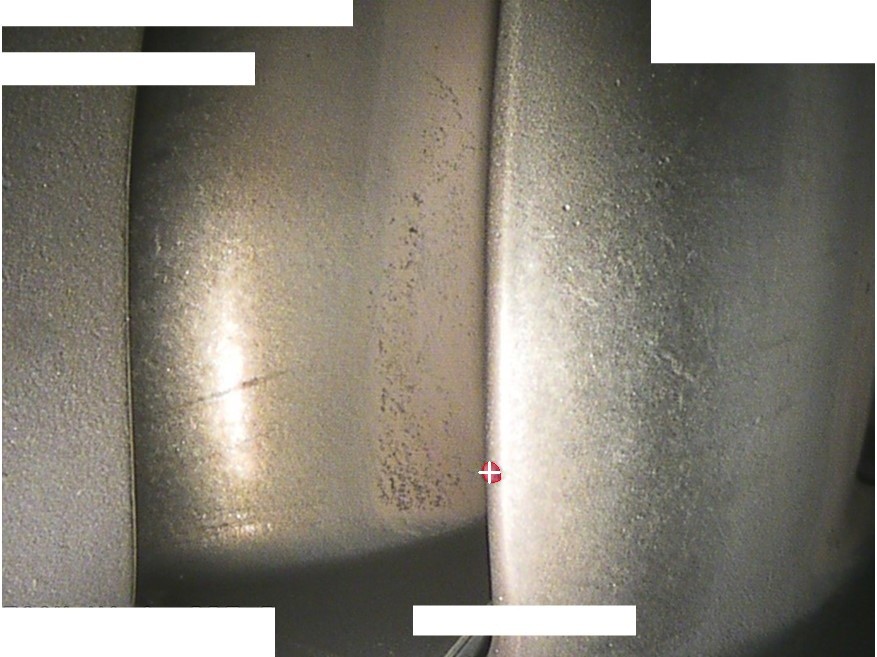}
        \end{subfigure}
        \begin{subfigure}[t]{0.195\linewidth}
            \caption{Missing material}
            \includegraphics[width=\linewidth]{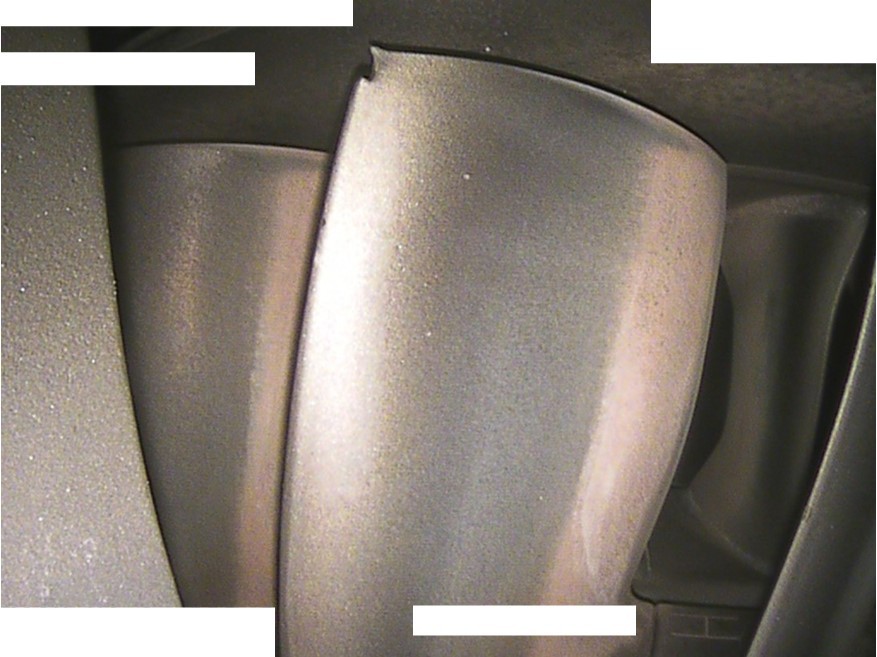}
            \includegraphics[width=\linewidth]{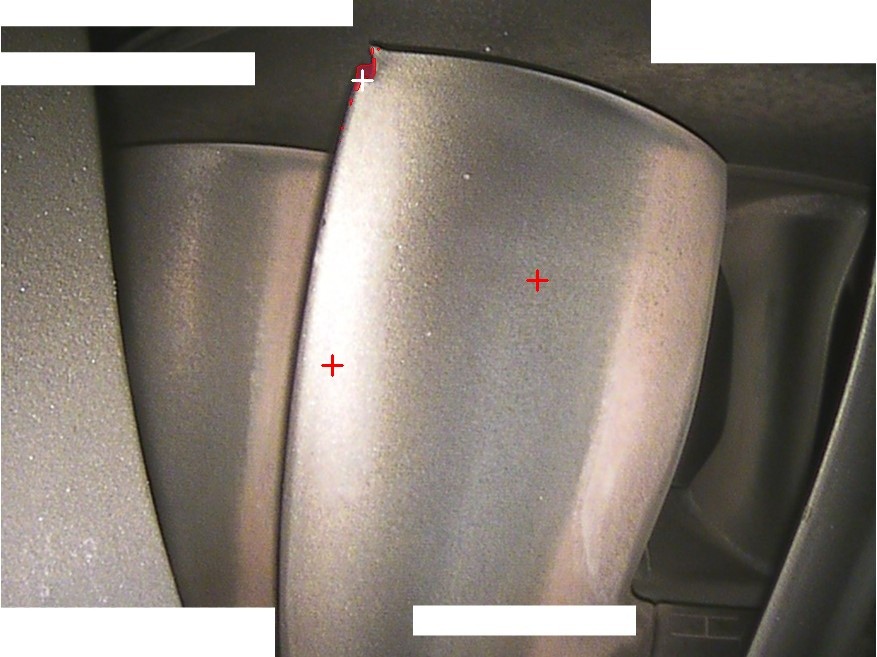}
        \end{subfigure}
    \end{subfigure}
    \caption{Borescope images \cite{li_deep_2022} from an aircraft engine for five defect classes: white and red crosses depict cursor inputs in background and foreground prompting (P.3).}
    \label{fig:BoroSamSamples}
\end{figure}

The annotation of image data for the inspection of low-contrast surfaces, such as those found in BI, is a tedious and time-consuming task. This is especially true when a pixel-by-pixel annotation is desired, e.g., in semantic segmentation tasks. SAM-based annotation drastically reduces the time required to annotate an image, as only a few prompts in the form of single cursor clicks are needed to generate a pixel-by-pixel ground truth segmentation map. Labeling tools such as AnyLabeling \cite{nguyen_anylabeling_2023} or Roboflow \cite{noauthor_roboflow_nodate} have already integrated the SAM into their annotation workflow. \Cref{fig:BoroSamSamples} shows the performance of the SAM in the inspection of aircraft engines with the endoscope. For five defect classes, only 1-3 cursor inputs by the user are required to create the segmentation map with pixel accuracy. 

In BI, in addition to the detection of a possible defect, the defect must also be measured. Depending on the defect class, the geometric extent, and the location of the defect, an appropriate measure is derived, such as taking the engine out of service.
Inspectors use a measuring borescope with integrated stereo cameras or a pattern projection to measure the defect’s surface, depth, length, or diameter.

\begin{figure}[!b]
	\centering
	\begin{subfigure}[t]{\linewidth}
		\begin{subfigure}[t]{0.32\linewidth}
			\caption{Separation of surface and noise}
			\includegraphics[width=\linewidth]{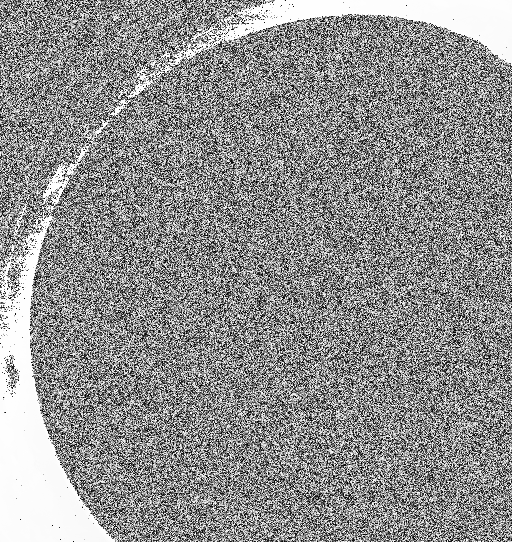}      
			\includegraphics[width=\linewidth]{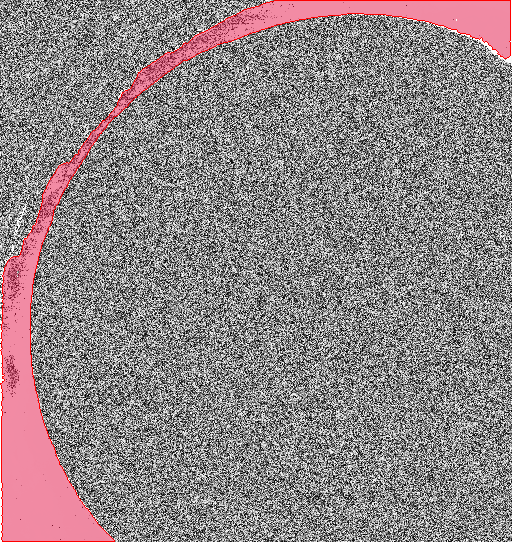}
		\end{subfigure}
		\begin{subfigure}[t]{0.32\linewidth}
			\caption{Anomalies}
			\includegraphics[width=\linewidth]{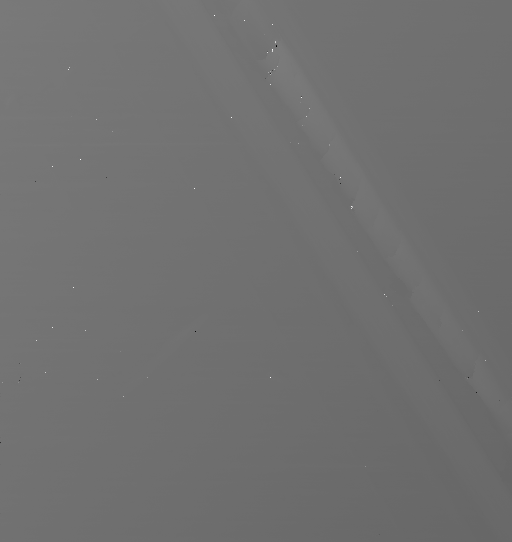}      
			\includegraphics[width=\linewidth]{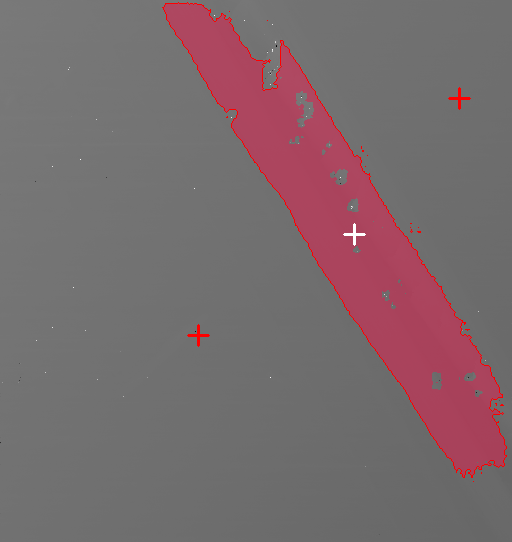}
		\end{subfigure}
		\begin{subfigure}[t]{0.32\linewidth}
			\caption{Scratch}
			\includegraphics[width=\linewidth]{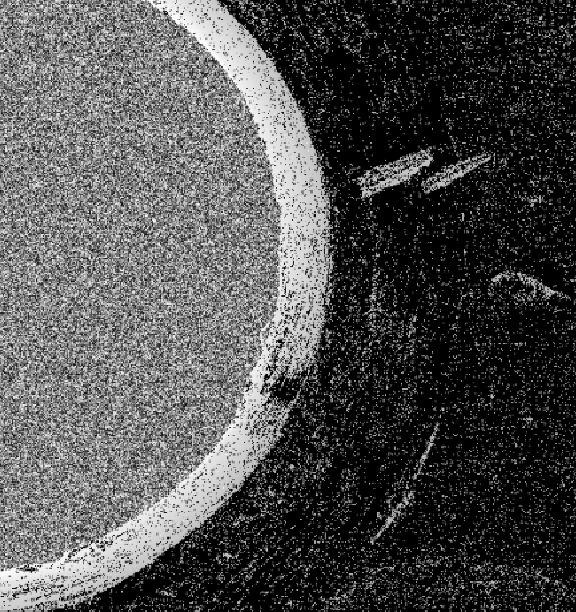}      
			\includegraphics[width=\linewidth]{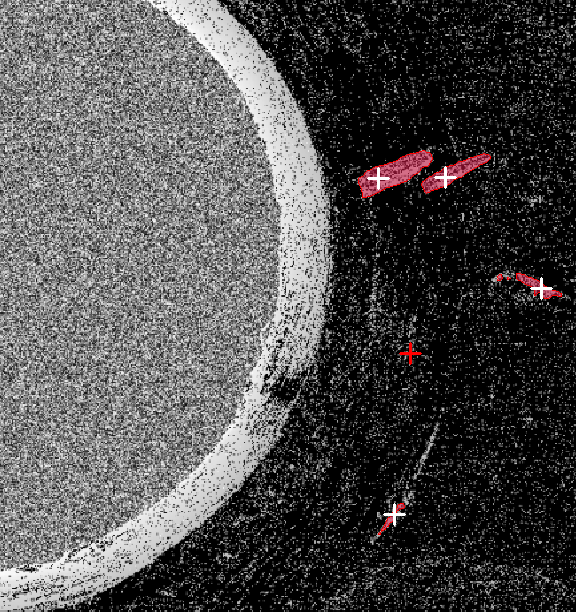}
		\end{subfigure}
	\end{subfigure}
	\caption{Application of the SAM to surface data of a White-Light Interferometer: (a) is generated through \textbf{P.1}. In (b,c) white and red crosses depict cursor inputs in foreground and background prompting (\textbf{P.3}), respectively.}
	\label{fig:WLI_segmentation}
\end{figure}

Measuring today is a tedious task because the effectiveness of the 3D measurement process on the stored 3D data depends heavily on manual effort. Manual cursor points are used as input in addition to a selected measurement type, such as length, depth, or surface. The integration of the SAM into the process promises to speed up the measurement time in the stored 3D data. If the defect in the image has been segmented with the SAM, the associated depth information, e.g., the length of a crack, can simply be subtracted from the 3D measurement. This simplifies the measurement because interfering 3D data that do not belong to the defect is removed.

\subsubsection{Non-Destructive Testing (NDT) based on White-Light Interferometry (WLI)}\label{sec:wli_inspection}

In the field of aircraft MRO, active research is being conducted on the integration of modern sensor technology into existing inspection technologies. White-light Interferometry (WLI) is a promising method in this regard. It is a non-destructive measuring principle that uses light paths of different lengths to analyze the surface of an object down to the nanometer range. In addition to parameters such as surface texture or roughness, damage such as dents, cracks, and chipping can be detected and analyzed.

Depending on the interferometer used, a large number of individual patches of a few mm\textsuperscript{2} each are recorded and registered together to form complete depth maps of object surfaces. The problem here is, on the one hand, the very high data density so that a high-resolution image of a surface segment can be generated and, on the other hand, a limited attention span of an inspector over a longer period of time. Assistance systems using the SAM can help either to recognize anomalies or to quickly select detected damage in the measurement data in order to be able to evaluate it using analytical procedures without also evaluating excess phenomena from the peripheral areas. 

Previous AI-supported methods in semantic segmentation, such as U-Net, require specific training data to achieve sufficient accuracy. For the sub-area of WLI outlined here, on the other hand, the necessary image quantities and high-quality annotations are not yet available. Consequently, the SAM offers the possibility of significantly reducing the problem of data availability.

\Cref{fig:WLI_segmentation} shows examples of various measurement recordings in which the SAM was used to segment areas from the recordings with just a few clicks and prepare them for further processing steps. For example, it is possible to separate measurement noise from surface information (a), select anomalies (b), or segment scratches (c).

\subsection{Process Monitoring and Control}\label{sec:mro_process_monitoring}%

\begin{figure}[b]
    \centering

    \begin{subfigure}[t]{0.49\linewidth}
        \includegraphics[width=\linewidth]{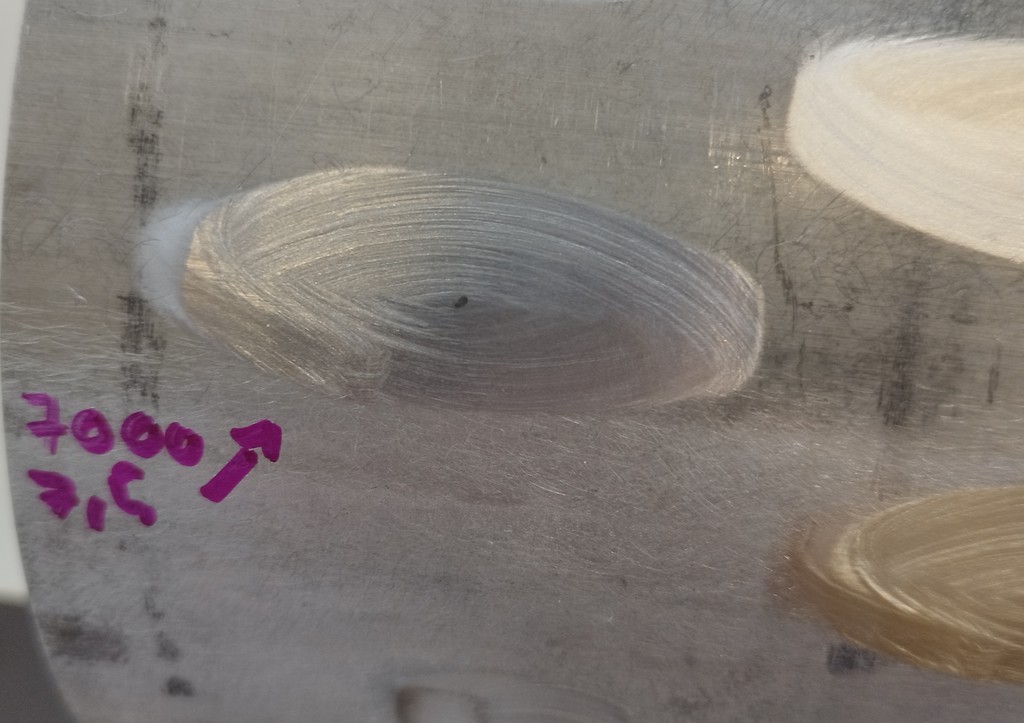}      
    \end{subfigure}
    \begin{subfigure}[t]{0.49\linewidth}
        \includegraphics[width=\linewidth]{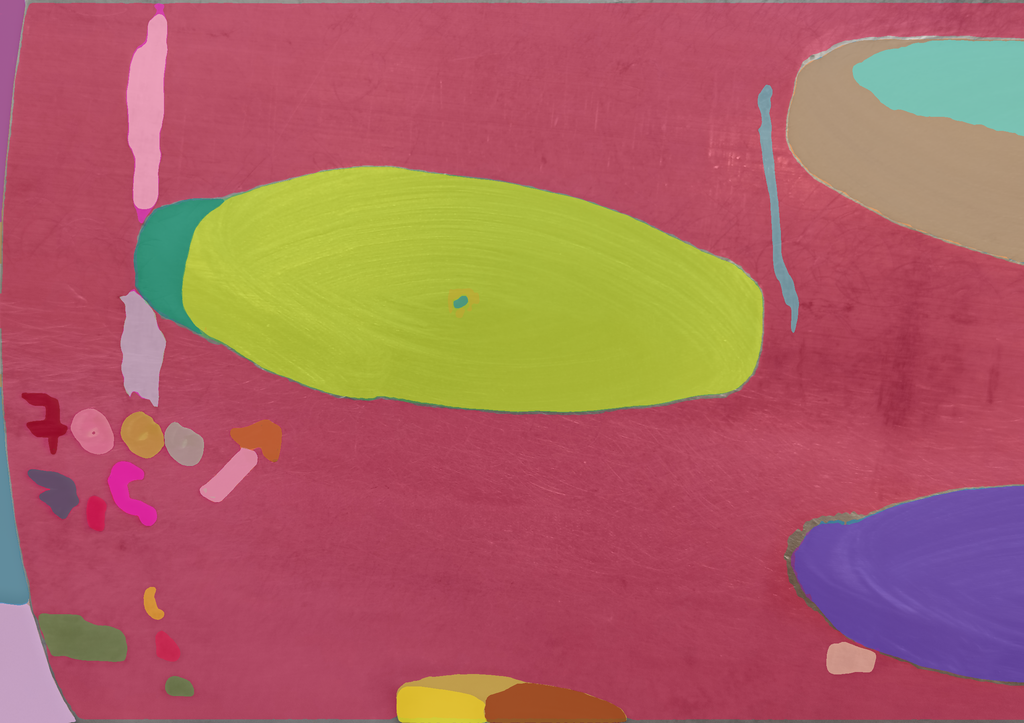}
    \end{subfigure}
    
    \caption{Intermediate inspection during surface defect removal by grinding.}
    \label{fig:defect_grinding_samples}
\end{figure}

Maintenance, repair, and overhaul of landing gear structures include visual inspection tasks performed by highly skilled personnel. Due to the complex component shapes, small corrosion defect extents, and difficult-to-see areas, visual inspection is a tedious and time-consuming task, inspection results are prone to error. As elaborated in \cref{sec:mro_visual_inspection}, using optical sensors and subsequent ai-based evaluation may relieve or assist the worker and lead to more reliable results, as well as easy documentation of defects and anomalies.

Identified defects have to subsequently be removed, which is usually performed manually by grinding. To maximize the chances of reusability of the component, the removal depth should be minimized \cite{kahlerAIbasedEndpointDetection2023}. However, as the defect depth cannot be determined reliably prior by means of NDT methods, an iterative procedure of grinding and checking for any left corrosion is pursued.
In order to increase the productivity of this repair process and to assist the worker or even automate the grinding process \cite{kahlerAutomatedGrindingSurface2023}, an optical sensor-based inspection of the grinding spot (rework) is proposed to detect the removal endpoint, requiring reliable detection of the defect in the developing grinding area.

Using the SAM segmentation capabilities, the approach would be segmenting the developing grinding area and the remaining defect within the grinding area, followed by a semantic decision to determine whether a defect is still present within the grinding spot segmentation. \Cref{fig:defect_grinding_samples} (a) shows a grinding spot with a defect still present. The elliptic grinding spot is evident. \Cref{fig:defect_grinding_samples} (b) shows the masks after automatic grid-based prompting (P.1). 

If a defect segment in a grinding spot is detected, more information could be obtained to adjust the grinding process and its parameters. For instance, the shape of the grinding point could be evaluated and compared with regulatory specifications. Moreover, the defect segment size, which is expected to decrease over the grinding iterations, could be calculated and used to predict the endpoint and avoid excessive material removal by adjusting process parameters such as grinding force, feed velocity, or grinding speed, as suggested in \cite{kahlerAIbasedEndpointDetection2023}.

\subsection{Segmenting 3D Scans of Unknown Environments}\label{sec:3D_scans}

\begin{figure}[!t]
    \centering
    \begin{subfigure}[t]{\linewidth}
        \centering
        \begin{subfigure}[t]{0.244\linewidth}
            \caption{2D camera-captured}
            \includegraphics[width=\linewidth]{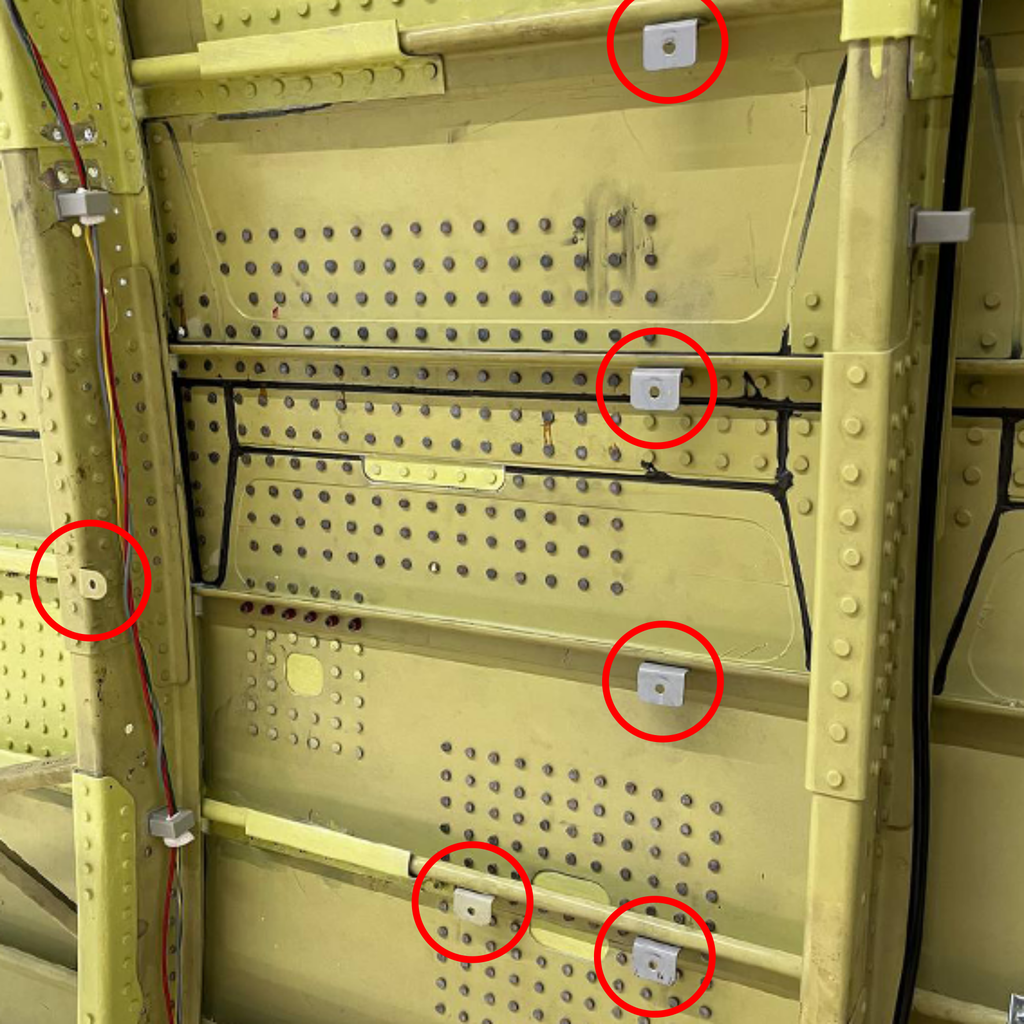}      
            \includegraphics[width=\linewidth]{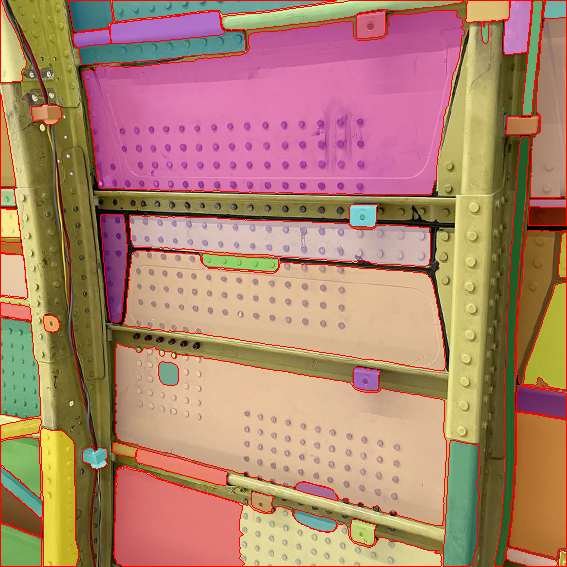}
        \end{subfigure}
        \begin{subfigure}[t]{0.244\linewidth}
            \caption{3D mesh (Artec Eva)}
            \includegraphics[width=\linewidth]{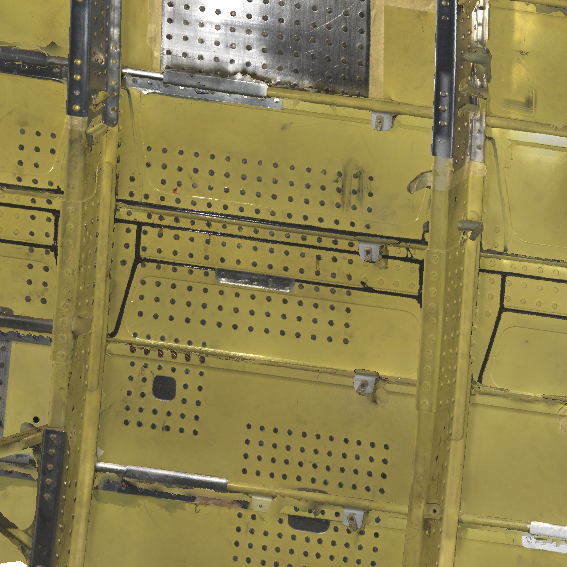}  
            \includegraphics[width=\linewidth]{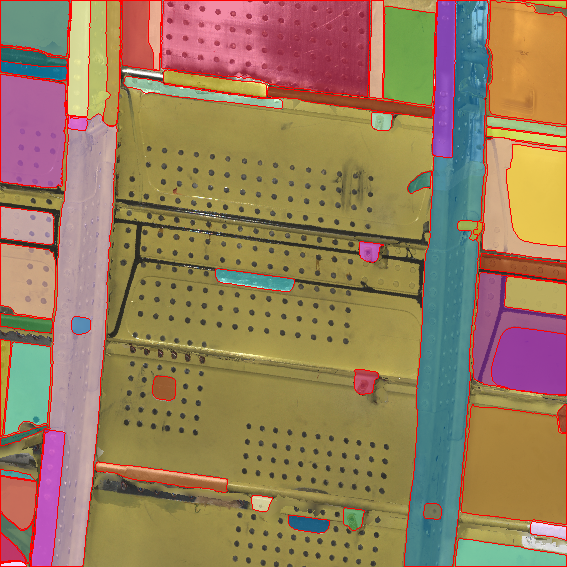}
        \end{subfigure}
        \begin{subfigure}[t]{0.244\linewidth}
            \caption{3D mesh (FARO Focus)}
            \includegraphics[width=\linewidth]{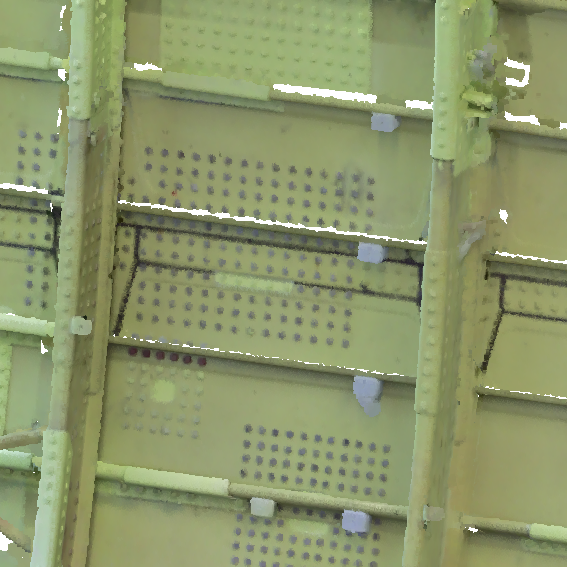}
            \includegraphics[width=\linewidth]{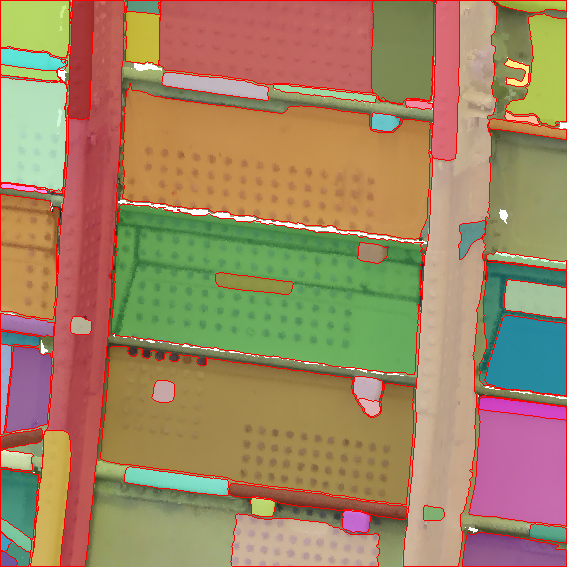}
        \end{subfigure}
        \begin{subfigure}[t]{0.244\linewidth}
            \caption{3D point cloud}
            \includegraphics[width=\linewidth]{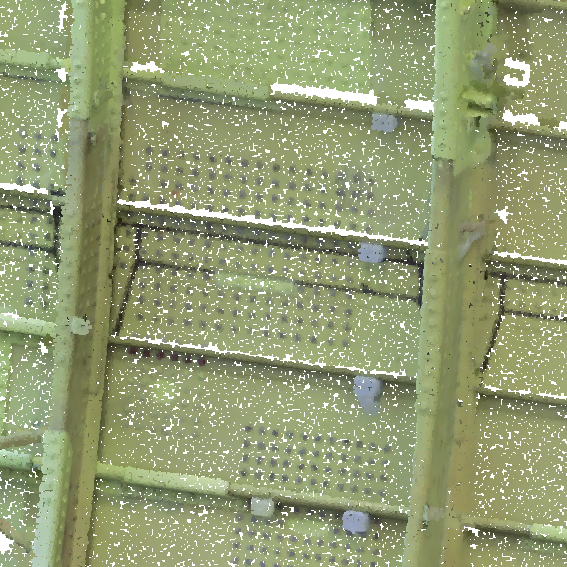}
            \includegraphics[width=\linewidth]{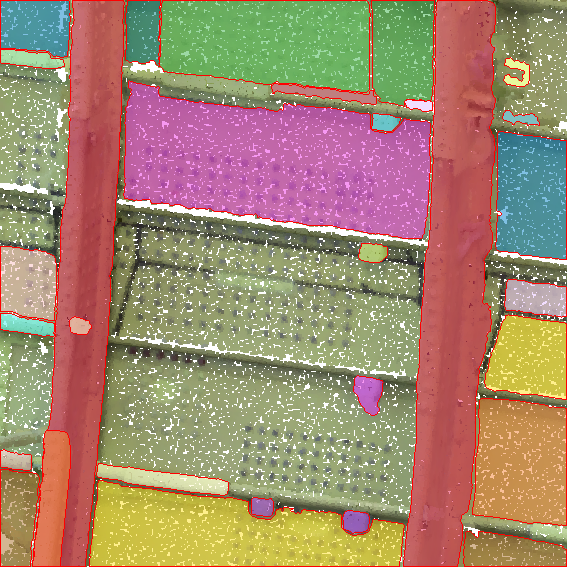}
        \end{subfigure}
    \end{subfigure}
    \caption{Automatic segmentation of an aircraft fuselage section captured in 2D and 3D (red circles around the objects of interest). Top row: input images; Bottom row: output.}
    \label{fig:fuselage_sample}
\end{figure}

The cabin is an expensive consumer product highly customized and regularly modified during the aircraft's lifespan. These modifications require an extensive planning phase based on a digital mock-up. The mock-up consists of 2D images, 3D scans, and CAD data \cite{moenck_towards_2022}. The data can be used for virtual fit checks, i.e., the positioning of space and allocation modules \cite{monchinger_automated_2021} and subsequent clash analysis \cite{laukotka_digitized_2019}, but also for spatially localizing objects of interest, e.g., brackets \cite{deneke_augmented_2021}. These tasks are usually done manually; an additional object layer on top of 3D data, even without semantics, can increase the engineering activities' time effectivity. When training a task-specific model, the challenge in this domain here is missing annotated training datasets, multi-modality, and recognition of texture-less, repetitive, and various different-sized objects. \cite{zhaoMaskRCNNBased2021} build a system to generate synthetic data of aircraft cable brackets in order to enable a deep learning task in an AR application for assembly inspection based on 2D acquired images \cite{hu_ar-based_2023, zheng_smart_2020}. Compared to 2D images, 3D-acquired data is usually of much lesser texture resolution but contains as-is spatial information.

Applying the SAM to first-hand acquired 3D data is already subject to different works: The most straightforward approach is to directly apply the SAM to the depth layer of an RGBD sensor, as shown in \cite{cenSADSegmentAny2023}, working surprisingly well, meaning the SAM can deal with texturelessness. Neural Radiance Fields (NeRFs) is a recent technique in view synthesis and 3D object reconstruction similar to photogrammetry working on multiple 2D images. Here, \cite{chenInteractiveSegmentAnything2023,cenSegmentAnything3D2023} already utilized the SAM to segment objects and parts in 3D. Differently, \cite{yang_sam3d_2023} projects textured point clouds from multiple viewpoints into 2D images and back-projects these masks again into the 3D space. The key step here is a proper merging approach since different viewpoints produce different per-point, non-semantic mask labels. Also, viewpoint-dependent occlusion must be taken into account. \cite{zhang_sam3d_2023} projects LiDAR data from a bird-eye perspective into a 2D image and applies the SAM to generate masks, also back-projecting these. What is interesting here is that the SAM is also working on these 2D-projected sparse point clouds. That also motivated us to test the SAM's robustness on 3D data in this work's depicted domain.

\begin{figure}[t]
    \centering
        \begin{subfigure}[t]{0.49\linewidth}
            \caption{Artec Eva}
            \includegraphics[width=\linewidth]{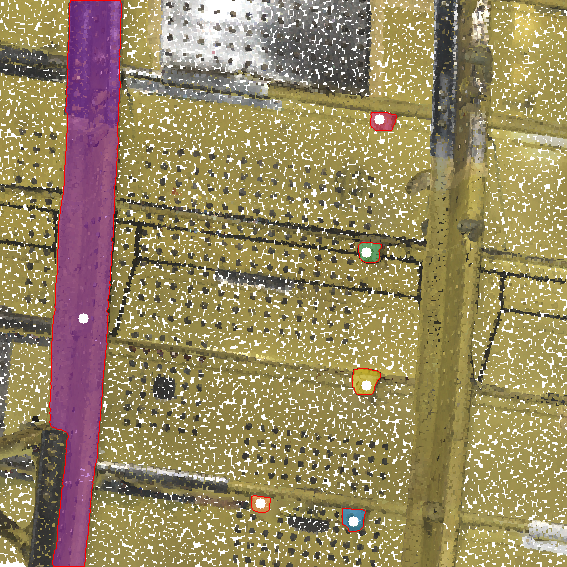}
        \end{subfigure}
        \begin{subfigure}[t]{0.49\linewidth}
            \caption{FARO Focus}
            \includegraphics[width=\linewidth]{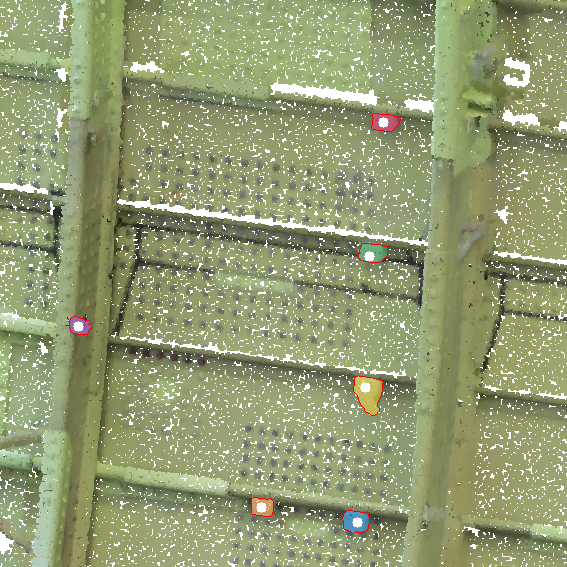}
        \end{subfigure}
    \caption{Single point prompting (P.2) the objects of interest in 3D-sampled point clouds.}
    \label{fig:fuselage_sparse_point_cloud_sample}
\end{figure}

\Cref{fig:fuselage_sample} shows the experimental use case, an aircraft fuselage section with six objects of interest, two different types of brackets (encircled in red in (a)). (a) images the section 2D camera-captured, while (b) and (c) are acquired by two different 3D scanning systems, Artec Eva and FARO Focus, respectively. While for (b) and (c), an orthographic projection into 2D of the textured mesh was fed into the SAM, for (d), a sparsely sampled point cloud ($0.1 \; points/mm^3$) was used after projection. \textbf{P.1} ($96$ points per side, $0.95$ pIoU, $0.5$ NMS threshold, $250$ minimum mask region area) was used for prompting, resulting in the bottom row that \cref{fig:fuselage_sample} shows. With the given parameters, the objects of interest, the brackets were successfully segmented while, e.g., smaller objects like rivet heads were omitted. Also, fewer masks are generated for the sampled point clouds. 
We also prompted the objects of interest by picking single points \textbf{P.2} in the sampled point clouds ($0.1 \; points/mm^3$). \Cref{fig:fuselage_sparse_point_cloud_sample} depicts the results. $11$ out of $12$ brackets were correctly segmented, while one time (s. (a) in \cref{fig:fuselage_sparse_point_cloud_sample}), the aircraft frame was masked. Fairly, the specific bracket is much lesser pronounced in texture in the specific 3D scan.

Up to date, in engineering activities during cabin overhaul, 3D scans of an aircraft cabin are used without additional point-wise information, mostly manually for, e.g., clash analysis or locating objects of interest. The SAM works surprisingly robust on objects in this domain. \cite{yang_sam3d_2023} presented an approach using the SAM for the segmentation of 3D point clouds; we showed that the SAM segments typical objects of interest in the 2D-projected space in the depicted domain. Merging both findings in a future application can definitely leverage different activities in a digital mock-up viewer or other applications. The SAM's robustness to concealed objects may help during manually surveying a cabin scan of millions of points to find an object of interest. In general, even non-semantic segmentation of otherwise less-textured 3D point clouds helps significantly during such an activity especially for non-expert operators. Also, the clash analysis can be conducted object-wise and not only point-wise. Besides, the SAM may be a component in any 3D semantic / instance / panoptic segmentation process in this domain.

\section{Summary}\label{sec:summary}

The results of this survey demonstrate the remarkable zero-shot capabilities of the SAM. All of the researchers involved in this study were astounded by its potential and how it could be applied to their individual use cases in an industrial / aircraft production-related context. Evaluation of the domain robustness definitely needs further investigation, especially to quantify the performance based on domain-related benchmark datasets. Task-related topics to further discuss are the type and parameters of prompting, the different levels of masks, and the split in generating an image embedding and prompting the model.

The promptable segmentation task can be utilized in generating pixel-level masks in 2D images in manual documentation, AR applications, and annotating training data. The zero-shot capabilities are so outstanding that basically in all of our use cases, intuitively clicking on a domain specific object or part worked as a human would expect.

According to our short survey and probably inexpressive view, the different levels of masks (whole, part, and sub-part) mostly worked in the context of defined clusterable objects. In the case of close-ups or other disturbances, e.g., WLI images (s. \cref{fig:WLI_segmentation}) or projecting point clouds (s. \cref{fig:fuselage_sparse_point_cloud_sample}), respectively, the mask with the highest predicted mIoU resulted in a meaningful semantic segment, while the rest of the masks were mostly cluttered, characterized by multiple non-connected regions. While it is not necessarily known a priori which of the three generated masks will be the best match for the desired application and prompting location, we observe that in most all cases, one of the masks does accurately match the expectation. Choosing a mask to use downstream in a processing pipeline is application dependent and, based on our observations, requires domain and context knowledge.

As outlined above, in our use cases, additional encoded or human-injected domain knowledge is required.
The most direct value created by the SAM can be observed in human-in-the-loop applications, as, e.g., found in \cref{sec:manual_quality_control}, where prompting location and mask can be selected by a user with domain knowledge. For non-interactive applications, the SAM must mostly be embedded as part of a processing pipeline, where mechanisms for generating prompting locations and mask selection have to be carefully chosen by embedding domain knowledge into the application. \Cref{fig:FoD_foam_inlay} shows a good example thereof, even for the case of automatic segmentation (\textbf{P.1}).

The novel concept introduced by the project of separating generating an image embedding and prompting the image is excellent in numerous applications besides annotating massive amounts of data in a web-based annotation tool. On the one hand, in applications in which multiple prompting is required; on the other hand, all applications in which the prompting might take place on a device not capable of holding the whole ViT in memory but must allow for prompting. Examples are applications running on embedded hardware, AR devices, or require a certain amount of human interactivity. However, image embedding generation must either be done remotely or using a scaled-down ViT, as demonstrated in MobileSAM \cite{zhang_faster_2023}.

\subsection{Industrial Applications}

Besides the variety of applications based on the SAM as shortly summarized in the related works \cref{sec:related_work}, in this use case study, we outlined the following production / industrial-related applications whose engineering efforts probably decrease while their robustness likely increases employing a VFM such as the SAM:

\begin{itemize}

    \item \textbf{Assembly supervision} (s. \cref{sec:assembly_progress_monitoring}) through progress monitoring and inspection is usually based on acquiring continuous frames imaging the same area or product. Using the SAM as an initial but also frame-to-frame mask generator is based on its zero-shot capabilities even possible in the depicted domain. For example, comparing the semantic-less masks between two frames outputs changes between two assembly steps. Similarly to comparing frames between two state, approaches like \cite{zhouDSECMOSSegmentAny2023,chengSegmentTrackAnything2023,zhangUVOSAMMaskfreeParadigm2023,yangTrackAnythingSegment2023} worked on the frame-by-frame segmentation in the context of videos.
    
    \item \textbf{Manual visual inspections within Quality Assurance} include, on the one hand, possible textual descriptions of found deviations but also often 2D images documenting regions of interest (s. \cref{sec:manual_quality_control}), often representing areas with visible surface defects. Generating even semantic-less masks accelerates but also increases the accuracy of these manual processes. The promptable segmentation tasks with single-point prompts intuitively support such human-executed visual inspection tasks.

    \item By merging human-generated prompts in 2D images with other modalities, like 3D scans, \textbf{measurement acquisition} can be supported. For example, measuring spatial characteristics of defects in 3D is a much more complex task than in 2D. Pixel-level masks in 2D transferred to 3D can accelerate such a process and assist human inspectors. Possible measuring systems are, e.g., borescopes.
    
    \item \textbf{Counting objects} play a role in many manufacturing tasks, especially in the aircraft industry, which deals with millions of different parts. Here, a VFM‘s zero-shot capabilities are specifically of interest. Counting is not directly enabled by the SAM but with adding minor additional logic as shown in \cite{maCanSAMCount2023a}. We identified that in our use cases (s. \cref{fig:logistics_use_cases}), the performance of this given approach is not sufficient enough for a real-world application in the depicted domain. However, it can be assumed that the vision community will continue working on this topic such that it is also applicable in the depicted domain.
    
    \item \textbf{Foreign Object Detection (FOD) and housekeeping} is a relevant topic in critical-regulated intralogistics domains like the aircraft industry. The given use cases in this study showed potential for automatic grid prompting (\textbf{P.1}) but also textual prompts (\textbf{P.5}). Especially applying textual prompts would decrease the level of necessary data and application engineering activities.
        
    \item \textbf{Bin picking} is a core problem in computer vision since a robot faces chaotic-ordered objects. Such a human-intuitive task requires massive engineering effort in a vision-based system. Pixel-accurate masks of unknown objects generated by the SAM can be used directly in a model-free grasp candidate generation pipeline or by fusing the 2D image mask with depth information. The SAM's zero-shot capabilities and mask hierarchy from part to the whole object are promising even facing small and cluttered objects, as shown in our use case (s. \cref{fig:logistics_use_cases}).
    
    \item Automated \textbf{anomaly detection in non-destructive inspection} processes are necessary when dealing with huge amounts of data. Humans are not able to conduct these inspections for more than a short period of time with the same repeatability and high objectivity. Facing anomalies, zero-shot capabilities are specific of interest. In two of the given use cases (s. \cref{fig:BoroSamSamples} and \cref{fig:WLI_segmentation}) mostly pixel-level accurate masked could be generated with the SAM.
       
    \item \textbf{In-process monitoring and control} using a cheap segmentation module like the SAM is a promising application as shown in \cref{sec:mro_process_monitoring}. The process changes between two consecutive frames, as in the example of local grinding, can be derived by prompting at the same position resulting in masks representing the same machined surface and vanishing defect.
    
    \item \textbf{Autonomous systems} require accurate representations of their surroundings to plan tasks and interact with their environment at all times. The segmentation capabilities of the SAM show great potential to aid in the necessary automated and assisted data creation through increased speed and accuracy of created maps and environmental models.
    Possible applications do not only include initial map creation but environment analysis, navigation, obstacle avoidance, and accurate segmentation for improved pose estimation and subsequent handling and manipulation.
    
\end{itemize}

\section{Discussion and Outlook}\label{sec:discss_outlook}

Aircraft as high-value, long-living, highly regulated assets result from complex production processes and depend on recurring Maintenance, Repair, and Overhaul (MRO) but also retrofit activities throughout the lifecycle. Deep learning-based machine vision in this field is based on the segregated character of the domain's objects not widely established since extensive training data is missing. Synthetic data and few-shot approaches are one research direction. However, the recent upstream of foundation models, in conjunction with strong zero-shot capabilities, lead in different areas to promising results in reducing the number of domain-specific datasets and also task-specific vision applications. In this work, we contributed here by surveying and recommending different industrial-characterized applications of the the SAM in aircraft-specific use cases from current research within the fields of manufacturing, intralogistics, and MRO.

Vision Foundation Models (VFM) like the SAM are rather novel but with increasing effort pursued by the community. 
Developing applications in an end-to-end manner is for unique and distinct domains data engineering-laborious.
VFMs usually only have a semantic understanding of "every day" or "world-known" objects.
The SAM demonstrated at first strong zero-shot but semantic-less capabilities even in segregated domains. It has learned some form of generic concepts of objects and parts.
So in vision applications that depend on semantics, at first, segmenting the scene into individual objects and subsequent processing with domain-specific knowledge may be more robust, time-efficient, and more economical in future vision application engineering processes.
So the idea is first to incorporate the knowledge learned by a foundation model and salting it with task and domain-specific knowledge in a further step.

Applying the SAM to different industrial real-world applications, as shown in this work, underlines its generalization capabilities. The initial analysis of the SAM's performance in the domain of aircraft production-related activities has shown promising results. Segmentation of objects has worked outstandingly; all of the authors encourage and demand further studies for VFM in industrial real-world applications, transferring the most recent progress from the vision community with the objective of benefiting industrial efficiency.

\newpage
\section*{Acknowledgment}
The research leading to these results received funding under the following Grant Numbers:
{
    \setlength\columnsep{60pt}
    \begin{multicols}{2}
    Federal Ministry for Economic Affairs and Climate Action (BMWK):
    \begin{itemize}
        \item 20D1916D - \emph{ADAPT}
        \item 20D1902C - \emph{InDiCaT}
        \item 20D2123E - \emph{ProDigieS}
        \item 20X1732B - \emph{MoSeBe}
        \item 20Q2109D - \emph{MEMS-Boro}
        \item 20Q1947B - \emph{WLIBoro}
    \end{itemize}
    
    \columnbreak
    
    Free and Hanseatic City Hamburg and IFB Hamburg:
    \begin{itemize}
        \item 51161730 - \emph{ILIdenT}
        \item 51165619 - \emph{prepAir 2}
        \item 51165621 - \emph{prepAir 3}
    \end{itemize}
    \end{multicols}
}

\section*{Authors' contributions}
All authors contributed to the work.
K. Moenck wrote the primary draft of the manuscript with the help of P. Pr{\"u}nte, D. Schoepflin, and A. Wendt.
K. Moenck, A. Wendt, P. Pr{\"u}nte, J. Koch, A. Sahrhage, J. Gierecker, O. Schmedemann, F. K{\"a}hler, and D. Holst contributed the use cases.
K. Moenck, A. Wendt, P. Pr{\"u}nte, and J. Koch conducted the final commenting and review process.
Funding was acquired by T. Sch{\"u}ppstuhl and M. Gomse.

\bibliography{bib_zotero_slice_0,bib_zotero_slice_1,bib_zotero_slice_2}

\end{document}